\documentclass[journal, final]{IEEEtran}
\usepackage{amssymb}
\usepackage{soul}
\usepackage{array}
\usepackage{amsfonts}
\usepackage{amsmath}
\usepackage{amsthm}
\usepackage{cases}
\usepackage{nccbbb}
\usepackage{epsfig}
\usepackage{epstopdf}
\usepackage{booktabs}
\usepackage{graphicx,subfigure}
\usepackage[ruled, linesnumbered]{algorithm2e}
\usepackage{url}
\usepackage{color}
\usepackage{multirow}
\usepackage{multicol}
\usepackage{makecell}
\usepackage{threeparttable}
\usepackage{booktabs}
\usepackage[caption=false,font=normalsize,labelfont=rm,textfont=rm]{subfig}
\usepackage[numbers,sort&compress]{natbib}
\usepackage{lineno}
\usepackage{rotating}
\usepackage{float}
\usepackage{bm}
\usepackage{pgfplots}
\pgfplotsset{compat=1.18}
 
\ifCLASSINFOpdf
\else
\fi

\usepackage[bookmarksnumbered,colorlinks,citecolor=blue,linkcolor=blue,urlcolor=black]{hyperref}
\makeatletter
\providecommand{\sf@counterlist}{}
\makeatother

\title{Adaptive Physical-Facial Representation Fusion via Subject-Invariant Cross-Modal Prompt Tuning for Video-Based Emotion Recognition}

\author{Xiwen Luo,
       Jia Li,
       Rencheng Song,~\IEEEmembership{Senior Member,~IEEE,}
       Yu Liu,~\IEEEmembership{Senior Member,~IEEE,}
       \\and Juan Cheng,~\IEEEmembership{Member,~IEEE}%
\thanks{X. Luo, R. Song, Y. Liu, and J. Cheng are with the Department
of Biomedical Engineering, and Anhui Province Key Laboratory of Measuring Theory and Precision Instrument, Hefei University of Technology, Hefei,
230009 China (e-mail: luoxiwen@mail.hfut.edu.cn; rcsong@hfut.edu.cn; yuliu@hfut.edu.cn; chengjuan@hfut.edu.cn).}%
\thanks{J. Li is with the School of Computer Science and Information Engineering, Hefei University of Technology, Hefei, 230601 China (e-mail: jiali@hfut.edu.cn).}%
\thanks{\textit{Corresponding authors: Jia Li, Juan Cheng.}}
}

\begin{document}
\maketitle
\begin{abstract}
Emotion recognition from facial videos enables non-contact inference of human emotional states. Facial expressions are the most widely used cues, but they cannot fully capture intrinsic affective states. Remote photoplethysmography (rPPG) offers a promising non-contact physiological cue, but it is susceptible to noise and pronounced inter-subject variability, which limits robust generalization to unseen individuals. Existing studies have integrated facial-expression cues with rPPG, yet current fusion mechanisms do not sufficiently address how physiological information should be incorporated without disrupting pretrained facial-expression knowledge, limiting the effective exploitation of cross-modal complementarity. Moreover, they lack explicit mechanisms to suppress subject-specific variations in rPPG signals, undermining cross-subject generalization. To address these issues, we propose a subject-invariant cross-modal prompt-tuning framework in which modality-complementary prompter (MCP) performs token-level prompt-guided interaction between rPPG and facial representations, while decoupled shared-specific adapter (DSSA) learns subject-invariant emotional representations. Specifically, we convert rPPG waveforms into noise-robust time--frequency representations (TFRs) and generate modality-complementary prompts to modulate facial tokens within a frozen Vision Transformer (ViT) backbone, thereby enabling cross-modal complementary interaction while maintaining generalizable facial representations. To further alleviate the influence of inter-subject shifts, we insert DSSAs into each ViT layer, explicitly decoupling subject-shared and subject-specific components to promote cross-subject generalization. Experiments on the MAHNOB-HCI and DEAP benchmarks show that the proposed method improves recognition accuracy and generalization ability over strong baselines, demonstrating its potential for video-based emotion recognition. The source code will be available at \url{https://github.com/MSA-LMC/SCPT}.
\end{abstract}



\section{Introduction}
\IEEEPARstart {E}{motion} is a psychological state influenced by both external stimuli and internal cognitive processes, which significantly influences human cognition, decision-making, and social interactions \cite{lerner2015emotion}. Automatically and accurately identifying emotional states (i.e., emotion recognition (ER) \cite{picard2000affective}) is crucial for developing empathetic artificial intelligence. It has been increasingly applied in human-computer interaction (HCI) fields such as intelligent driving, telemedicine, and online education.

Existing ER data sources can be broadly categorized into peripheral behavioral cues (e.g., facial expressions (FEs), speech, and body movements) and physiological signals (e.g., electrocardiography (ECG), photoplethysmography (PPG) and electrodermal activity). Compared with peripheral behavioral cues, physiological signals reflect intrinsic affective states that are difficult to consciously regulate \cite{xu2025hypercomplex, bamonte2023emotion}. Nevertheless, their acquisition typically requires contact-based sensors, restricting their use in universal applications or real-world environments. The emergence of remote photoplethysmography (rPPG) \cite{song2020heart, yu2022physformer, cheng2023motion, liu2024rppg, zou2025rhythmmamba} offers a viable non-contact alternative by analyzing subtle chromatic fluctuations in facial videos induced by cardiac activity to extract remote blood volume pulse (rBVP) signals. These signals enable the estimation of physiological indicators such as heart rate (HR) and heart rate variability (HRV), serving as effective markers of emotion \cite{benezeth2018remote, yu2019remote, mellouk2023cnn}. However, rPPG-derived physiological patterns exhibit pronounced inter-subject variability, which makes robust cross-subject emotion recognition challenging, especially with limited training data.

\begin{figure}[t]
	\centering
	\includegraphics[width=\linewidth]{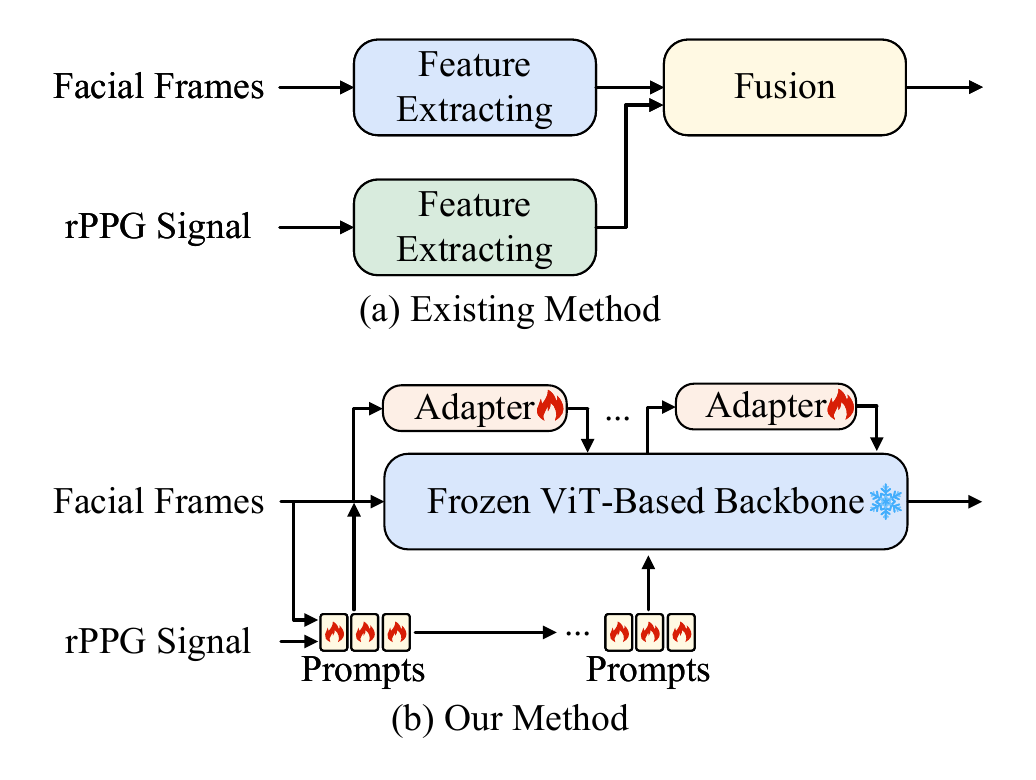}
	\caption{Schematic comparison between common rPPG-facial emotion recognition paradigms (a) and the proposed framework (b). In our framework (b), modality-complementary prompters (MCPs) inject rPPG-derived prompts into a pretrained facial-expression backbone to introduce complementary physiological cues while preserving facial emotion semantics, whereas decoupled shared-specific adapters (DSSAs) promote subject-invariant emotional representation learning during fine-tuning.}
	\label{fig:intro_compare}
\end{figure}

Leveraging the abundant in-the-wild labels and well-established priors in the facial expression recognition (FER) domain, recent studies integrate FE and rPPG signals as complementary modalities to enhance emotion recognition accuracy \cite{ouzar2022video, wu2023recognizing, li2024end}. Ouzar et al. \cite{ouzar2022video} first demonstrated that combining rPPG with FE features substantially improved recognition accuracy over unimodal approaches. Wu et al. \cite{wu2023recognizing} integrated rPPG-derived physiological measures, including remote HRV (rHRV), with fine-grained facial descriptors for complex emotion classification. To further exploit the potential correlations between rPPG and facial features, Li et al. \cite{li2024end} employed transformer-based cross-modal attention mechanism to construct an end-to-end multimodal emotion recognition (MER) framework. Bao et al. \cite{bao2025svd} further proposed an SVD-guided hierarchical multimodal fusion method that combines facial expression, rPPG, and gaze cues, using contrastive and consistency constraints to suppress fusion redundancy and enhance discriminative emotional representations. However, existing fusion mechanisms still do not sufficiently address how physiological information should be integrated without disrupting pretrained facial-expression knowledge. When modalities are fused in a roughly equal or unconstrained manner, the relatively weak and noisy rPPG cues may interfere with the facial semantics learned from FER priors, thereby limiting the effective exploitation of cross-modal complementarity \cite{bao2025svd}, as schematically summarized in Fig.~\ref{fig:intro_compare}(a).

In addition, inter-subject variability is a well-known challenge in emotion recognition, causing notable performance degradation under cross-subject evaluation \cite{shen2023contrastive}. To mitigate this issue, recent studies have explored feature disentanglement-based domain generalization methods \cite{wang2023generalizing} that explicitly separate subject-shared factors from subject-specific nuisance factors \cite{zhao2021plug, jia2024multi, lyu2025mi, yang2026fddgnet, gao2025multimodal, piratla2020efficient, li2018deep}. However, these methods typically depend on heavily engineered auxiliary objectives or additional training branches (e.g., adversarial learning \cite{zhou2025eegmatch} or reconstruction constraints \cite{li2025emotion}), which are brittle in the domain-generalization setting with limited multimodal data. Moreover, imperfect disentanglement may either suppress subtle affective cues or still preserve spurious subject-specific artifacts, thereby limiting robustness.

To tackle the above challenges, we propose SCPT, a \textbf{s}ubject-invariant \textbf{c}ross-modal \textbf{p}rompt-\textbf{t}uning framework (see Fig.~\ref{fig:intro_compare} (b)). In SCPT, modality-complementary prompters (MCPs) inject rPPG-derived prompts into a pretrained facial-expression backbone to introduce complementary physiological cues while preserving facial emotion semantics, whereas decoupled shared-specific adapters (DSSAs) promote subject-invariant emotional representation learning during fine-tuning. Specifically, we first convert rPPG waveforms into noise-robust time--frequency representations (TFRs) and extract fine-grained physiological features using lightweight residual CNN blocks. These features are then used by MCPs to generate rPPG-derived prompts, which are injected into Vision Transformer (ViT) layers of the pretrained facial-expression backbone to introduce complementary physiological cues while preserving facial emotion semantics. Besides, to further mitigate the influence of inter-subject shifts, DSSAs are inserted into each ViT layer to decouple subject-shared and subject-specific components, thereby progressively refining multimodal representations toward subject-invariant emotion encoding. As a result, the proposed SCPT learns subject-generalizable emotional representations from limited training data, leading to improved robustness. Extensive experiments on MAHNOB-HCI and DEAP benchmarks demonstrate that the proposed method consistently outperforms strong baselines in both accuracy and generalization. The main contributions of this work are summarized as follows:

\begin{enumerate}
  \item A subject-invariant cross-modal prompt-tuning (SCPT) framework for video-based emotion recognition is proposed, in which modality-complementary prompters (MCPs) fuse rPPG features from time-frequency representations (TFRs) with facial features to generate prompts for a frozen ViT-based facial-expression backbone, thereby facilitating cross-modal complementary interaction to improve emotion recognition accuracy.

  \item To improve robustness under inter-subject variability, decoupled shared-specific adapters (DSSAs) are inserted into each ViT layer to perform subject-invariant correction through explicit feature decoupling of subject-shared and subject-specific components, thereby enhancing the model's ability to accommodate individual variability and improve generalization.

  \item Comprehensive experiments on MAHNOB-HCI and DEAP datasets demonstrate the proposed SCPT's superior recognition accuracy and generalization across subjects, providing a universal foundation for robust non-contact multimodal emotion recognition.
\end{enumerate}

\section{Related Work}

\subsection{Emotion Recognition with rPPG Signals}
Physiological signals such as ECG and PPG have been shown to exhibit strong correlations with human emotional states \cite{ismail2022comparison, sarkar2020self}, but their contact-based sensors limit their use in many real-world settings. Recently, rPPG \cite{song2020heart, yu2022physformer, cheng2023motion, liu2024rppg, zou2025rhythmmamba} has emerged as a non-contact alternative that estimates rBVP from subtle skin color variations in facial videos, enabling physiological emotion recognition without physical sensors.

Benezeth et al. \cite{benezeth2018remote} first demonstrated that rHRV features extracted from facial chrominance signals can serve as effective physiological indicators of emotion. Building upon this finding, Yu et al. \cite{yu2019remote} developed a deep spatiotemporal network to reconstruct high-fidelity rPPG signals and used rHRV features for emotion classification via SVM. Mellouk et al. \cite{mellouk2023cnn} further applied independent component analysis (ICA) to extract pulse-wave components from facial RGB signals and utilized a framework consisting of 1D-CNN and LSTM for continuous emotion estimation. Despite the growing utility of rPPG in ER, several critical challenges persist. The inherent low amplitude of rPPG signals, coupled with their high susceptibility to noise from illumination variations and motion artifacts, substantially degrades the accuracy of ER \cite{bao2025svd}. Moreover, the discriminative ability of rPPG alone is not consistently reliable across different affective dimensions, especially under cross-subject variability and limited training data. In contrast, facial expressions provide direct behavioral cues reflecting external affective states \cite{zhu2024dynamic}. Integrating rPPG with facial-expression cues enables joint modeling of autonomic physiology and appearance-based behavior, improving robustness to noisy sensing and motivating multimodal fusion for more reliable affect recognition \cite{ouzar2022video, wu2023recognizing, li2024end}.

\begin{figure*}[t]
	\centering
	\includegraphics[width=1.0\linewidth]{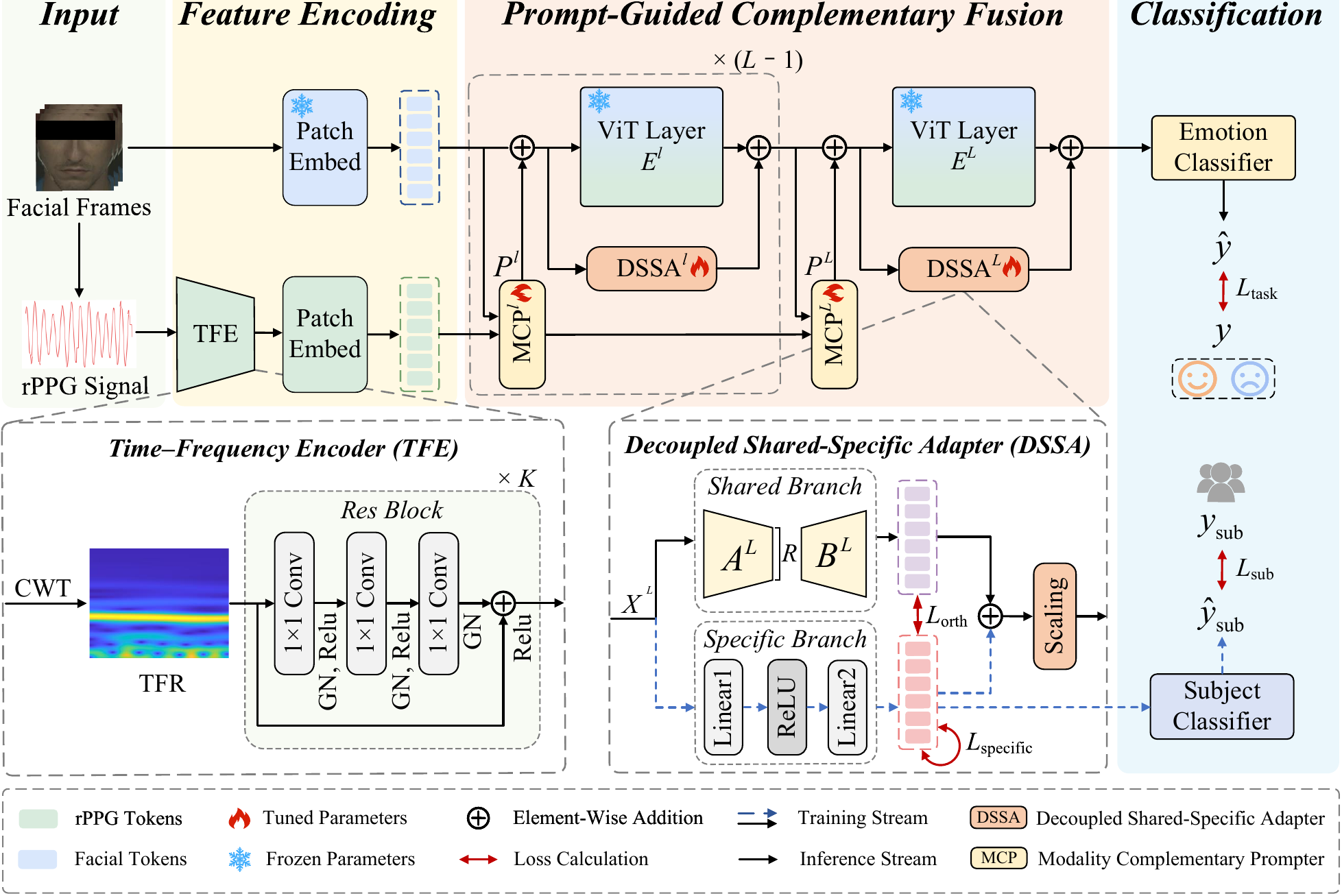}
\caption{Overall architecture of the proposed SCPT. Facial frames and rPPG time-frequency representation (TFR), where the latter are obtained from rPPG signals via continuous wavelet transform (CWT), are first fed into the patch embedding layers to generate corresponding facial and physiological tokens. Fine-grained physiological features are then injected into the visual token space to form modality-complementary prompts (MCPs), which guide the fine-tuning of the stacked pretrained ViT backbone and introduce complementary physiological cues while preserving facial emotion semantics. During model adaptation, the parameters of the proposed decoupled shared-specific adapters (DSSAs) inserted into each layer remain trainable to perform subject-invariant feature correction, facilitating collaborative cross-modal modeling and improved cross-subject generalization.}
	\label{fig:architecture}
\end{figure*}

\subsection{Emotion Recognition with FE and rPPG Signals}
Facial expressions and rPPG signals are respectively emotion-related behavioral and physiological cues extracted from facial videos, each with inherent complementary properties in emotion representation. Ouzar et al. \cite{ouzar2022video} showed that combining rPPG with FE features improves recognition performance over using a single modality. Wu et al. \cite{wu2023recognizing} further combined HRV-based physiological descriptors with facial features, leading to more accurate classification of complex emotions. However, these studies mainly rely on coarse fusion and do not consider the latent complementarity between facial expressions and physiological signals. Recently, Li et al. \cite{li2024end} proposed a Transformer-based cross-modal attention framework to capture the complementary relationship between rPPG and facial features, creating a robust end-to-end multimodal emotion recognition system. Bao et al. \cite{bao2025svd} introduced an SVD-guided hierarchical multimodal fusion strategy that combines facial expression, rPPG, and gaze cues to reduce redundancy and enhance discriminative emotion representations. Nevertheless, existing methods still lack an effective mechanism for fine-grained, semantics-preserving integration of rPPG-derived physiological cues into facial-expression representations. Coarse fusion or unconstrained feature mixing can interfere with pretrained facial affect semantics and underuse weak pulse-related evidence \cite{bao2025svd}. Consequently, cross-modal complementarity is often not reliably translated into consistent recognition gains. Moreover, they do not explicitly account for the pronounced inter-subject variability of physiological cues, which limits generalization to unseen individuals. Therefore, this work aims to propose a prompt-guided fusion of facial expressions and rPPG to inject complementary physiological cues into a pretrained facial backbone, enabling subject-invariant representations for improved recognition and cross-subject generalization.

\subsection{Disentanglement-Based Cross-Subject Emotion Recognition}
Inter-subject variability has long been recognized as a major obstacle to generalizable emotion recognition \cite{wang2023generalizing}. To address this issue, disentanglement-based domain generalization (DG) methods have been widely explored, aiming to decouple emotion-relevant representations from subject-specific nuisance factors at the feature level \cite{zhao2021plug, jia2024multi, lyu2025mi, yang2026fddgnet}. The core idea behind these approaches is to decompose learned representations into a shared component that captures emotion-discriminative information invariant across subjects and a private component that models individual-specific variations \cite{piratla2020efficient, li2018deep}. Han et al. \cite{han2023noise} proposed a framework that disentangles raw electroencephalography (EEG) signals into subject/session-specific features, motor imagery task-specific features, and random noise, improving robustness and interpretability. Wu et al. \cite{wu2024grop} introduced a graph orthogonal purification network that enhances individual adaptability by promoting orthogonality and transferability between emotion-relevant and emotion-irrelevant features. Lyu et al. \cite{lyu2025mi} proposed a mutual-information disentanglement module with prototype-based class constraints to separate domain-related and domain-unrelated features and improve cross-domain semantic consistency. Yang et al. \cite{yang2026fddgnet} introduce an orthogonal complementary projection and an information-bottleneck-inspired constraint to separate emotion-relevant and subject-specific subspaces while preserving reconstruction fidelity. Despite their effectiveness, these methods often rely on carefully designed disentanglement objectives and auxiliary constraints, which are sensitive to optimization instability and limited training data \cite{zhao2021plug,yang2026fddgnet}. Moreover, imperfect disentanglement may either attenuate subtle emotion-related patterns or still retain spurious subject-specific artifacts. These limitations are further amplified in multimodal settings involving weak and noisy physiological signals, such as rPPG, where robust disentanglement and generalization remain challenging.

\section{Method}
This section presents our SCPT framework for video-based emotion recognition. As illustrated in Fig.~\ref{fig:architecture}, the overall architecture consists of three modules: a feature encoding module, a prompt-guided complementary fusion module, and a classification module. The feature encoding module introduces stable rPPG TFRs to extract robust physiological features. The prompt-guided complementary fusion module, built on a pretrained ViT backbone, enables cross-modal interaction and complementary fusion via MCP, followed by representation refinement using DSSA. The classification module produces the final emotion predictions. The following subsections detail the key designs of these components: Stable rPPG Time-Frequency Encoding, Prompt-Guided Complementary Fusion, Decoupled Shared-Specific Adaptation, and Emotion Classification and Overall Optimization.

\subsection{Stable rPPG Time-Frequency Encoding}
\begin{figure}[htbp]
	\centering
	\includegraphics[width=\linewidth]{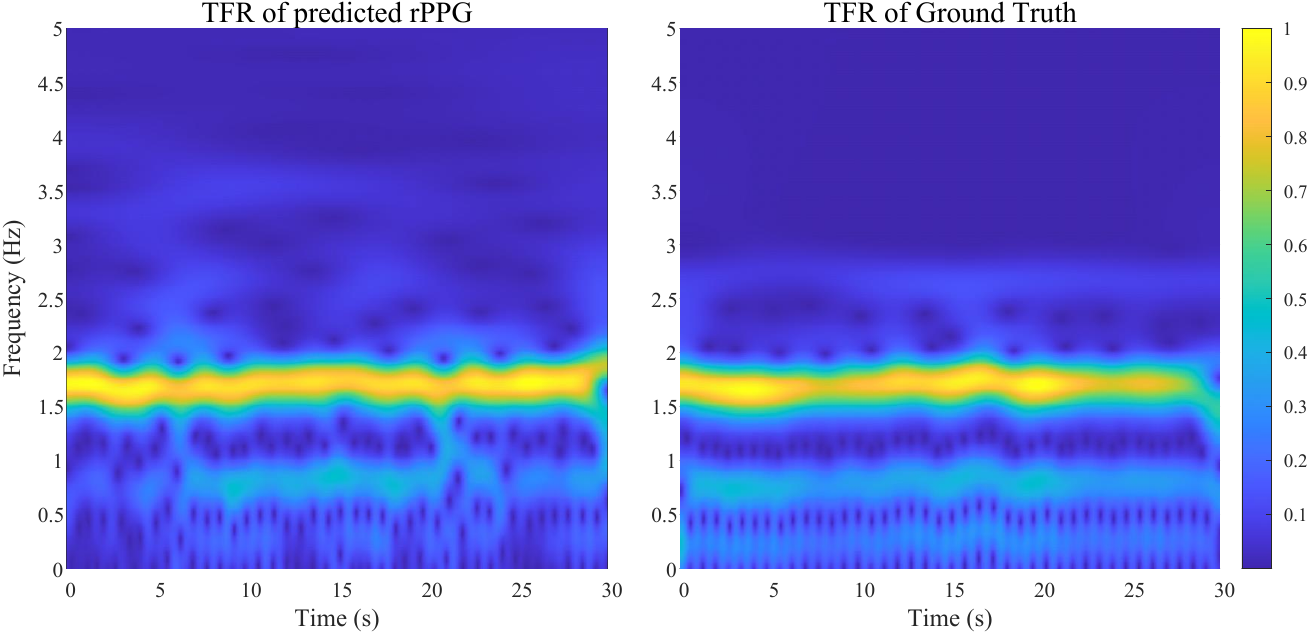}
	\caption{Comparison of the TFRs of the predicted rPPG (left) and the reference PPG (right) from a 30-s DEAP segment. In both plots, the dominant physiological rhythm appears as a bright horizontal ridge in the low-frequency region. Compared with the reference PPG, the predicted-rPPG TFR preserves this main band but is slightly broader and less smooth, with more diffuse background energy, indicating residual noise and motion artifacts.}
	\label{fig:tfr_comparison}
\end{figure}

Benefiting from recent advances in high-accuracy rBVP recovery, facial videos can provide reliable rPPG signals as physiological inputs. Specifically, the rPPG waveform is extracted from raw facial video using RhythmMamba \cite{zou2025rhythmmamba}. However, it is a variable time-domain signal that is easily affected by subject differences, motion, illumination changes, and recording conditions, making direct fusion with facial features unstable. To address this issue, we introduce a time--frequency encoder (TFE) \cite{cheng2026video}, which converts the waveform into a 2D time–frequency representation (TFR) $X_{\mathrm{TFR}}$. This representation preserves rhythmic physiological structure mitigating noise-induced variations, thereby facilitating subsequent cross-modal interaction. As illustrated in Fig.~\ref{fig:tfr_comparison}, the extracted rPPG TFR preserves the dominant physiological rhythm of the reference PPG, i.e., the simultaneously recorded contact-based ground-truth signal, while suppressing noise and artifacts in a more structured form for subsequent cross-modal interaction. The TFR is then fed into the physiological encoder $g(\cdot)$ to obtain $X_{\mathrm{R}}=g(X_{\mathrm{TFR}})$ for subsequent multimodal modeling.

\subsection{Prompt-Guided Complementary Fusion}

To fully leverage the abundant prior information provided by facial expressions, we adopt a Vision Transformer (ViT) pretrained on the large-scale static facial expression dataset AffectNet \cite{mollahosseini2017affectnet} as the backbone and directly utilize the released pretrained weights from Chen et al. \cite{chen2024static}. For each video clip, $T$ frames are uniformly and randomly sampled as input, denoted as $X_{\mathrm{F}} \in \mathbb{R}^{T \times C \times H \times W}$, where $T$, $C$, $H$, and $W$ denote the number of frames, channels, height, and width of the facial image sequence, respectively. The $T$ frames are first rearranged into the batch dimension, so that each frame is processed independently by the subsequent modules. Each frame is then partitioned into $N$ local patches by a patch embedding layer and mapped into a $D$-dimensional latent space, yielding $H_{\mathrm{F}}^0 \in \mathbb{R}^{N \times D}$ per frame. These patch embeddings, along with a [class] token $x_{\mathrm{class}}$ and positional encoding $X_{\mathrm{pos}} \in \mathbb{R}^{(N+1) \times D}$, are then fed into the Transformer encoder layers $\{E^l\}_{l=1}^{L}$.

To achieve effective cross-modal interaction and complementary information fusion between the facial and rPPG representations, we adopt a multi-view complementary prompter (MCP) \cite{zhu2023visual}. The rPPG time-frequency features are first transformed into token representations $H_{\mathrm{R}}^0 \in \mathbb{R}^{N \times D}$ per frame via a patch embedding layer. Following Zhu et al. \cite{zhu2023visual}, MCP projects facial and physiological tokens into a shared latent space and generates modality-complementary prompts through attention-guided interaction. These prompts are then added to the visual token sequence to inject physiological information into facial representations, enabling prompt-guided cross-modal interaction and complementary fusion between facial and rPPG features.

The prompts for the $l$-th Transformer layer are given by $P^{l} \in \mathbb{R}^{N \times D}$, which stacks $N$ prompt tokens of dimension $D$, and are dynamically generated based on the feature representation and prompts from the previous layer as
\begin{equation}
P^{l} = \mathrm{MCP}^{l}\left(H_{\mathrm{F}}^{\,l-1},\,P^{\,l-1}\right),\quad l=1,\ldots,L,
\end{equation}
where $H^0 = H_{\mathrm{F}}^0$ and $P^0 = H_{\mathrm{R}}^0$. The learned prompts $P^l$ are then injected into the current visual token sequence in a residual manner as
\begin{align}
(H^{\,l-1})' &= H^{\,l-1} + P^{\,l}, \\
H^{\,l} &= E^{\,l}\Big(\big[(x_{\mathrm{class}},(H^{\,l-1})')\big] + X_{\mathrm{pos}}\Big),
\end{align}
where $H^{\,l}$ denotes the updated visual token representation after the $l$-th layer. This residual injection mechanism enables MCP to continuously perform cross-modal semantic interaction and complementary information enhancement throughout the Transformer hierarchy.

\subsection{Decoupled Shared-Specific Adaptation}
In multimodal emotion recognition, inter-subject variability often leads to severe performance degradation on unseen subjects \cite{jia2024multi}. Recent work achieves adaptation to user-specific domain shifts with only a few unlabeled samples through parameter-efficient fine-tuning \cite{hsieh2026alfa}. Inspired by this, we propose a Decoupled Shared-Specific Adapter (DSSA) to model subject-shared patterns and subject-dependent nuisance variations separately under limited adaptation capacity, while preserving the frozen backbone semantics.

The proposed DSSA is attached to the frozen backbone in a parallel residual manner. For the $l$-th Transformer layer, we denote the input token sequence to the backbone by $X^{\,l}=(H^{\,l-1})'$. Below, we describe the DSSA model structure.

\paragraph{Shared-Specific Feature Decoupling}
For each Transformer layer $l$, the feature correction is decomposed as
\begin{equation}
\Gamma^{\,l}(X^{\,l}) = \Gamma_{\mathrm{shared}}^{\,l}(X^{\,l}) + \Gamma_{\mathrm{specific}}^{\,l}(X^{\,l}).
\label{eq:shared_subspace}
\end{equation}
Here, $\Gamma_{\mathrm{shared}}^{\,l}(X^{\,l}) = X^{\,l}A^{\,l}(B^{\,l})^\top$, $A^{\,l}, B^{\,l} \in \mathbb{R}^{D \times R}$ parameterize a shared low-rank subspace with $R \ll D$. This shared branch is intentionally implemented in a LoRA-style form, since subject-shared patterns are expected to lie in a compact and transferable subspace and should be modeled with limited adaptation capacity. Conversely, $\Gamma_{\mathrm{specific}}^{\,l}(\cdot)$ is implemented as a lightweight two-layer MLP. This specific branch adopts a standard nonlinear adapter form, since subject-specific deviations and acquisition-related noise are typically more heterogeneous and residual in nature.

At each Transformer block, DSSA employs a parallel residual connection, where the adapted representation is computed as
\begin{equation}
H^{\,l} = E^{\,l}\Big(\big[(x_{\mathrm{class}},X^{\,l})\big] + x_{\mathrm{pos}}\Big) + s \cdot \Gamma^{\,l}(X^{\,l}),
\label{eq:forward_parallel}
\end{equation}
where $s$ is a residual scaling factor. 

\paragraph{Emotion-Relevant Subspace Analysis}
Although the low-rank component is designed to capture subject-invariant structure, it still absorbs task-irrelevant variations that are broadly shared across samples, such as illumination-induced background fluctuations and camera or codec distortions. Therefore, we perform a subspace analysis on the learned shared representations to extract the dominant emotion-relevant directions. Given the final-layer shared features $\Gamma_{\mathrm{shared}}^{\,L}(X^{\,L}) \in \mathbb{R}^{N \times D}$, we compute a rank-$S$ truncated singular value decomposition (SVD) to extract the emotion-relevant component as

\begin{equation}
\Gamma_{\mathrm{shared}}^{\,L}(X^{\,L}) \approx U_s \Sigma_s V_s^\top,
\label{eq:vemo_derivation}
\end{equation}
where $U_s \in \mathbb{R}^{N \times S}$, $V_s \in \mathbb{R}^{D \times S}$, and $\Sigma_s \in \mathbb{R}^{S \times S}$ are the rank-$S$ left singular matrix, right singular matrix, and diagonal singular-value matrix, respectively. Since SVD ranks feature-space patterns by energy, the leading components captures the dominant variation structure of high-dimensional representations. Accordingly, retaining the leading $S$ components preserves the main structure of shared representations, which contains the most discriminative emotion-related information while reducing redundancy. Similar strategies have been adopted in related research fields \cite{bao2025svd, hsieh2026alfa, wei2001ecg}.

\subsection{Emotion Classification and Overall Optimization}
Finally, we obtain the predicted emotion-class probability distribution $\hat{y}$ via the emotion classifier $C_{\mathrm{emo}}$ as
\begin{equation}
\hat{y} = C_{\mathrm{emo}}\!\big(\text{CLS}(X^{\,L})\,V_s\big),
\label{eq:emo_projection_cls}
\end{equation}
where $\text{CLS}(X^{\,L}) \in \mathbb{R}^{1 \times D}$ extracts the $[\text{class}]$ row from the model output features $X^{\,L}$, and $V_s$ stacks the $S$ leading right singular directions from \eqref{eq:vemo_derivation}, which span the dominant emotion-relevant subspace in feature space.

The task loss is defined as $L_{\mathrm{task}} = \text{CrossEntropy}(\hat{y}, y)$, where $y$ denotes the discrete emotion label. Additionally, to enforce a clear separation between identity and emotion factors in DSSA while optimizing the emotion-recognition task, we apply three complementary constraints. First, to prevent the specific branch from overfitting to random fluctuations or leaking affective cues, we apply an $L_1$ sparsity penalty as
\begin{equation}
L_{\mathrm{specific}} = \frac{1}{L}\sum_{l=1}^{L} \| \Gamma_{\mathrm{specific}}^{\,l}(X^{\,l}) \|_1.
\label{eq:specific}
\end{equation}
Second, we reduce information overlap between the shared and specific branches with an orthogonality loss
\begin{equation}
L_{\mathrm{orth}} = \frac{1}{L}\sum_{l=1}^{L} \|\Gamma_{\mathrm{shared}}^{\,l}(X^{\,l})(\Gamma_{\mathrm{specific}}^{\,l}(X^{\,l}))^\top\|_F^2.
\label{eq:orth}
\end{equation}
Third, we add subject-discriminative supervision by feeding the subject-specific features into an auxiliary classifier $C_{\mathrm{sub}}$ so that they explicitly encode identity-related information, with the corresponding loss defined as
\begin{equation}
L_{\mathrm{sub}} = \text{CrossEntropy}\!\Big(C_{\mathrm{sub}}\big(\text{CLS}(\Gamma_{\mathrm{specific}}^{\,L}(X^{\,L}))\big), y_{\mathrm{sub}}\Big),
\label{eq:sub}
\end{equation}
where $\Gamma_{\mathrm{specific}}^{\,L}(X^{\,L})$ denotes the subject-specific branch output at the last Transformer layer, $\text{CLS}(\cdot)$ selects the CLS token feature used for subject classification, and $y_{\mathrm{sub}}$ denotes the ground-truth subject identity label.
The complete training objective integrates the task loss, sparsity regularization, orthogonality regularization, and subject supervision, which is written as
\begin{equation}
L_{\mathrm{total}} = L_{\mathrm{task}} + \lambda_1 L_{\mathrm{specific}} + \lambda_2 L_{\mathrm{orth}} + \lambda_3 L_{\mathrm{sub}}.
\label{eq:loss_dss}
\end{equation}

During training, the trainable modules (the rPPG encoder, patch embedding, MCP, DSSAs, and the classifiers) are updated by minimizing $L_{\mathrm{total}}$, while the pretrained ViT backbone remains frozen.
During inference on unseen subjects, the specific branch can be deactivated to perform subject-invariant inference, ensuring that the model relies solely on subject-invariant affective cues.

\section{Experiments and Results}

\begin{table*}[htbp]
  \centering
  \caption{Comparative experimental results on the MAHNOB-HCI dataset. Accuracy (\%) is reported in mean$\pm$std format. CAP and CA denote direct concatenation-projection fusion and cross-attention fusion, respectively. The best results are in bold.}
  \label{tab:mahnob_results}
  \resizebox{0.95\textwidth}{!}{
  \begin{tabular}{l l l c c}
  \toprule
  \multirow{2}{*}{Studies} & \multirow{2}{*}{Modalities} & \multirow{2}{*}{Features} & \multicolumn{2}{c}{ACC} \\
  \cmidrule(lr){4-5}
  & & & Valence & Arousal \\
  \midrule
  Ferdinando et al. \cite{ferdinando2017emotion} (2018) & ECG & HRV & 70.70$\pm$4.90 & 73.50$\pm$4.40 \\
  Oğuz et al. \cite{oguz2023emotion} (2023) & ECG & Morphological, HRV & 78.28 & 83.61 \\
  Singh et al. \cite{singh2023bio} (2023) & Bio-Sensing & Bio-sensing deep learning-based & 65.44$\pm$7.18 & 68.70$\pm$10.03 \\
  Jia et al. \cite{jia2024multi} (2024) & Bio-Sensing & Bio-sensing deep learning-based & 66.00 & 66.30 \\
  \midrule
  Siddharth et al. \cite{jung2019utilizing} (2019) & Facial, Bio-Sensing & Face appearance, PSD, Physiological image-based etc. & 85.49 & 82.93 \\
  Li et al. \cite{li2021mindlink} (2021) & Facial, EEG & Face appearance, PSD & 78.56 & 77.22 \\
  Tan et al. \cite{tan2021neurosense} (2021) & Facial, EEG & Facial landmarks, EEG deep learning-based & 72.12 & 79.39 \\
  Zhu et al. \cite{zhu2024dynamic} (2024) & Facial, EEG & Face appearance, PSD & 79.37$\pm$9.00 & 75.62$\pm$10.14 \\
  Ali et al. \cite{ali2025unified} (2025) & Facial, Bio-Sensing & Face appearance, Physiological image-based & 50.01 & 83.84 \\
  Wu et al. \cite{wu2026dhcm} (2026) & Facial, EEG & Face appearance, EEG deep learning-based & 75.12$\pm$9.43 & 73.78$\pm$11.16 \\
  \midrule
  Mellouk et al. \cite{mellouk2023cnn} (2023) & rPPG & rPPG deep learning-based & 73.33 & 60.00 \\
  Li et al. \cite{li2024end} (2024) & Facial, rPPG & Face appearance, rPPG deep learning-based & 68.54$\pm$5.86 & 66.83$\pm$13.54 \\
  CAP (Ours) & Facial, rPPG & Face appearance, rPPG deep learning-based & 71.67$\pm$8.36 & 66.67$\pm$10.58 \\
  CA (Ours) & Facial, rPPG & Face appearance, rPPG deep learning-based & 69.17$\pm$6.38 & 70.83$\pm$11.46 \\
  SCPT (Ours) & Facial, rPPG & Face appearance, rPPG deep learning-based & \textbf{75.83$\pm$5.24} & \textbf{74.17$\pm$9.85} \\
  \bottomrule
  \end{tabular}
  }
  \end{table*}

\subsection{Datasets}
\textbf{MAHNOB-HCI} \cite{soleymani2011multimodal} contains multimodal recordings from 30 participants, including facial videos, EEG, ECG, respiration amplitude, skin temperature, and audio signals. All physiological data are precisely synchronized with the corresponding facial video recordings. Each participant provided self-assessment ratings of valence, arousal, and dominance, which reflect their spontaneous emotional responses, using a 9-point scale (from 1 to 9). As the data from three participants were corrupted, the experiments were conducted on the remaining 27 subjects, resulting in a total of 527 complete video recordings for analysis.

\textbf{DEAP} \cite{koelstra2012deap} consists of recordings from 32 participants, each watching 40 one-minute music videos. It provides synchronized physiological signals, including 32-channel EEG, PPG, galvanic skin response (GSR), and respiration signals, all sampled at 128 Hz after preprocessing. Participants rated each video on valence, arousal, dominance, and liking, using a 9-point Likert scale. Facial videos were recorded for 22 out of the 32 participants, and only these 22 subjects were included in our analysis.

\subsection{Data Pre-processing}
In data preprocessing stage, MediaPipe \cite{lugaresi2019mediapipe} is employed to detect and crop facial regions from video frames, after which the cropped face images are resized to $224 \times 224$. Following Chen et al. \cite{chen2024static}, 16 RGB frames are uniformly sampled from each facial video clip as model input. For rPPG signal extraction, we employ the RhythmMamba network trained on the VIPL-HR and UBFC-rPPG datasets to extract rPPG signals from raw facial videos. During training, RhythmMamba uses simultaneously recorded ECG or photoplethysmography (PPG) signals as ground-truth (GT). In the MAHNOB-HCI dataset, the EXG2 channel from the three-lead ECG system is used as GT, while in the DEAP dataset, the PPG signal is used as GT. The GT signals are filtered and normalized following Ali et al. \cite{ali2025unified}. Following Jung et al. \cite{jung2019utilizing}, we apply Continuous Wavelet Transform (CWT) with the Analytic Morse Wavelet to compute the TFR and focus on the 0--5 Hz frequency band. Before being fed into the model, the rPPG TFRs are resized to $224 \times 224$. We use a three-layer ResNet \cite{he2016identity} as the physiological encoder $g(\cdot)$ to process the rPPG TFRs.

The same data pre-processing strategy is applied to these two datasets, among which the last 30-s of each trial of MAHNOB-HCI dataset is extracted as the analysis sample after removing the self-rating period based on timestamps \cite{li2024end, zhang2022gcb, zhu2024dynamic}. To alleviate the issue of limited sample size, following Ali et al. \cite{ali2025unified}, each 30-second trial was further segmented into 5-second non-overlapping clips, resulting in multiple samples per trial. Importantly, all clips derived from the same trial share the same emotion label. Consistent with prior works, we binarized the valence and arousal scores using a threshold of 5, thereby converting them into two-class discrete labels. The classification performance was evaluated separately for valence and arousal recognition tasks.

\subsection{Experiment Settings and Implementation Details}
To verify the effectiveness and generalization capability of the proposed method, subject-independent leave-one-subject-out (LOSO) \cite{jung2019utilizing, singh2023bio} evaluation was conducted on both the MAHNOB-HCI and DEAP datasets. Each subject was used as a test fold once, with the remaining data divided 80/20 into training and validation sets. The reported mean and standard deviation are computed across all test subjects under the protocol. For the training strategy, the AdamW optimizer was employed with a base learning rate of $1 \times 10^{-4}$ and a weight decay of 0.01. A cosine annealing schedule was adopted to gradually reduce the learning rate over 100 epochs. During training, only the rPPG time-frequency feature extraction residual blocks, the patch embedding layer, the MCP, DSSA parameters, and the classifiers were trainable, while all other parameters remained frozen. Key hyperparameters and dataset-specific auxiliary loss weights are summarized in Table~\ref{tab:param_settings} (see Fig.~\ref{fig:lambda_sensitivity} for $\lambda_i$ sensitivity). All experiments were implemented in PyTorch and conducted on two NVIDIA RTX A6000 GPUs.

\begin{table}[htbp]
\renewcommand{\arraystretch}{1.12}
\centering
\caption{Parameter settings on two datasets. Auxiliary weights $(\lambda_1,\lambda_2,\lambda_3)$ refer to \eqref{eq:loss_dss} and are chosen near the peaks in Fig.~\ref{fig:lambda_sensitivity}. Valence and Arousal denote separate binary classification heads trained with the corresponding rows.}
\label{tab:param_settings}
\begin{tabular}{@{}lcc@{}}
\toprule
Settings & MAHNOB-HCI & DEAP \\
\midrule
Optimizer & AdamW & AdamW \\
Batch size & 32 & 64 \\
Base learning rate & $1{\times}10^{-4}$ & $1{\times}10^{-4}$ \\
Weight decay & 0.01 & 0.01 \\
LR schedule & Cosine (100 epochs) & Cosine (100 epochs) \\
DSSA $R$ / $s$ / $S$ & 8 / 0.1 / 16 & 8 / 0.1 / 32 \\
$(\lambda_1,\lambda_2,\lambda_3)$ (Valence) & (0.2, 0.1, 0.6) & (0.3, 0.3, 0.8) \\
$(\lambda_1,\lambda_2,\lambda_3)$ (Arousal) & (0.1, 0.1, 0.6) & (0.3, 0.3, 0.4) \\
\bottomrule
\end{tabular}
\end{table}

\subsection{Comparison with Existing Methods}
We evaluate the proposed model on two widely used benchmarks, MAHNOB-HCI and DEAP, under the experimental protocol described above. The comparisons include representative baselines using facial cues, physiological cues, and their combinations. Emotion recognition performance is measured by Accuracy (ACC) and F1-Score. Quantitative results are reported in Tables \ref{tab:mahnob_results} and \ref{tab:deap_results}.

On MAHNOB-HCI, the proposed method achieves the strongest performance among the non-contact facial-video-based methods listed in Table \ref{tab:mahnob_results}. Relative to the unimodal rPPG model of Mellouk et al. \cite{mellouk2023cnn}, the proposed model improves Valence accuracy from 73.33\% to 75.83\% and Arousal accuracy from 60.00\% to 74.17\%. It also consistently surpasses the existing facial+rPPG baselines, including Li et al. \cite{li2024end}, direct concatenation-projection fusion (CAP), and cross-attention fusion (CA). These results indicate that jointly modeling facial appearance and rPPG through prompt-guided cross-modal interaction is more effective than direct concatenation and cross-attention fusion. Although some methods using contact biosignals such as EEG or ECG still achieve higher absolute accuracy, the proposed method remains competitive while relying only on non-contact facial video.

\begin{table*}[htbp]
\centering
\caption{Comparative experimental results on the DEAP dataset. Accuracy (\%) is reported in mean$\pm$std format. CAP and CA denote direct concatenation-projection fusion and cross-attention fusion, respectively. The best results are in bold.}
\label{tab:deap_results}
\resizebox{0.95\textwidth}{!}{
\begin{tabular}{l l l c c}
\toprule
\multirow{2}{*}{Studies} & \multirow{2}{*}{Modalities} & \multirow{2}{*}{Features} & \multicolumn{2}{c}{ACC} \\
\cmidrule(lr){4-5}
& & & Valence & Arousal \\
\midrule
Siddharth et al. \cite{jung2019utilizing} (2019) & Bio-sensing & PSD, Physiological image-based etc. & 71.87 & 73.05 \\
Elalamy et al. \cite{elalamy2021multi} (2021) & Bio-Sensing & Physiological image-based & 69.90 & 69.70 \\
Jia et al. \cite{jia2024multi} (2024) & Bio-Sensing & Bio-sensing deep learning-based & 61.50 & 65.30 \\
Li et al. \cite{li2025uncertainty} (2025) & Bio-Sensing & Bio-sensing deep learning-based & 69.62$\pm$7.32 & 70.62$\pm$7.96 \\
Tian et al. \cite{tian2026heterogeneity} (2026) & Bio-Sensing & Bio-sensing deep learning-based & 60.07$\pm$5.65 & 64.16$\pm$9.61 \\
\midrule
Siddharth et al. \cite{jung2019utilizing} (2019) & Facial, EEG & Face appearance, PSD & 79.52 & 78.34 \\
Tan et al. \cite{tan2021neurosense} (2021) & Facial, EEG & Facial landmarks, EEG deep learning-based & 67.76 & 78.97 \\
Zhu et al. \cite{zhu2024dynamic} (2024) & Facial, EEG & Face appearance, PSD & 72.22$\pm$6.88 & 70.69$\pm$8.94 \\
Gao et al. \cite{gao2025multimodal} (2025) & Facial, Bio-Sensing & Face appearance, EEG\&ECG deep learning-based & 65.84 & 64.62 \\
Wu et al. \cite{wu2026dhcm} (2026) & Facial, EEG & Face appearance, EEG deep learning-based & 72.89$\pm$6.38 & 71.71$\pm$8.86\\
\midrule
Wu et al. \cite{wu2023recognizing} (2023) & Facial, rPPG & Face appearance, rHRV deep learning-based & 60.00 & \textbf{72.50} \\
Li et al. \cite{li2024end} (2024) & Facial, rPPG & Face appearance, rPPG deep learning-based & 61.25$\pm$9.47 & 63.75$\pm$13.62 \\
CAP (Ours) & Facial, rPPG & Face appearance, rPPG deep learning-based & 63.75$\pm$11.18 & 62.92$\pm$10.86 \\
CA (Ours) & Facial, rPPG & Face appearance, rPPG deep learning-based & 62.50$\pm$10.26 & 61.25$\pm$12.58 \\
SCPT (Ours) & Facial, rPPG & Face appearance, rPPG deep learning-based & \textbf{67.92$\pm$8.45} & 65.41$\pm$9.73 \\
\bottomrule
\end{tabular}
}
\end{table*}

The results on DEAP further verify the effectiveness of the proposed method under a more challenging setting. Compared with MAHNOB-HCI, DEAP involves stronger inter-subject variation, longer affective stimuli, and less constrained recording conditions, which generally lead to lower overall performance. Even under these conditions, the proposed model achieves the best Valence result among all facial+rPPG methods, obtaining 67.92\% on Valence, while also delivering competitive Arousal performance at 65.41\%.

In particular, the proposed method outperforms Li et al. \cite{li2024end} by 6.67 percentage points on Valence and 1.66 percentage points on Arousal, and also improves over both direct concatenation-projection fusion (CAP) and cross-attention fusion (CA). These gains suggest that MCP and DSSA remain effective under stronger cross-subject variability and less controlled acquisition, where fine-grained prompt-based physiological injection and subject-invariant adaptation are particularly beneficial. Compared with Wu et al. \cite{wu2023recognizing}, the proposed method performs better on Valence but remains lower on Arousal. A plausible explanation is that the rHRV-based descriptors adopted by Wu et al. provide stronger direct physiological cues for arousal, whereas the proposed framework emphasizes unified end-to-end representation learning.

\section{Analysis and Discussion}

\subsection{Ablation Studies}
\subsubsection{Ablation on Key Components}
We quantify each module by progressively enabling the Time--Frequency Encoder (TFE), Multi-View Complementary Prompter (MCP), Decoupled Shared-Specific Adapter (DSSA), and the SVD-based emotion subspace regularization, as summarized in Tables~\ref{tab:component_ablation} and~\ref{tab:component_ablation_deap}. The configuration with TFE but without MCP reproduces the CAP baseline in Tables~\ref{tab:mahnob_results} and~\ref{tab:deap_results}; relative to disabling TFE (first row), enabling TFE alone does not yield consistent gains across metrics, indicating that a stable rPPG view by itself is insufficient under coarse multimodal coupling. Enabling MCP produces clearer improvements, especially on DEAP, suggesting that prompt-guided interaction better exploits cross-modal complementarity than coarse multimodal coupling alone, with larger gains when recording conditions are less constrained and cross-subject variability is stronger. DSSA further refines most metrics by subject-invariant adaptation, though the stepwise trend is not strictly monotonic on every score. The full configuration delivers the best overall balance on both benchmarks, indicating that time--frequency encoding, prompt-based interaction, shared-specific decoupling, and subspace guidance are complementary rather than redundant.

\begin{table}[htbp]
  \centering
  \caption{Ablation results on key components on MAHNOB-HCI. Accuracy and F1-Score (\%) are reported in mean$\pm$std format. The best results are in bold. The following tables are reported in the same format.}
  \label{tab:component_ablation}
  \resizebox{0.48\textwidth}{!}{
  \begin{tabular}{c@{\hspace{4pt}}c@{\hspace{4pt}}c@{\hspace{4pt}}c@{\hspace{6pt}}c c c c}
  \toprule
  \multirow{2}{*}{TFE} & \multirow{2}{*}{MCP} & \multirow{2}{*}{DSSA} & \multirow{2}{*}{SVD} & \multicolumn{2}{c}{Valence} & \multicolumn{2}{c}{Arousal} \\
  \cmidrule(lr){5-6} \cmidrule(lr){7-8}
   &  &  &  & ACC & F1 & ACC & F1 \\
  \midrule
    &   &   &   & 71.67$\pm$9.84 & 76.39$\pm$13.18 & 70.00$\pm$12.47 & 68.48$\pm$15.42 \\
  $\checkmark$ &   &   &   & 71.67$\pm$8.36 & 77.70$\pm$12.64 & 66.67$\pm$10.58 & 69.35$\pm$14.73 \\
  $\checkmark$ & $\checkmark$ &   &   & 75.00$\pm$8.27 & 74.14$\pm$9.96 & 71.67$\pm$11.46 & 70.21$\pm$11.83 \\
  $\checkmark$ & $\checkmark$ & $\checkmark$ &   & 75.42$\pm$6.92 & 76.28$\pm$13.57 & 73.33$\pm$12.68 & 71.84$\pm$13.71 \\
  $\checkmark$ & $\checkmark$ & $\checkmark$ & $\checkmark$ & \textbf{75.83$\pm$5.24} & \textbf{77.17$\pm$8.45} & \textbf{74.17$\pm$9.85} & \textbf{72.76$\pm$11.67} \\
  \bottomrule
  \end{tabular}
  }
\end{table}

\begin{table}[htbp]
  \centering
  \caption{Ablation results on key components on DEAP.}
  \label{tab:component_ablation_deap}
  \resizebox{0.48\textwidth}{!}{
  \begin{tabular}{c@{\hspace{4pt}}c@{\hspace{4pt}}c@{\hspace{4pt}}c@{\hspace{6pt}}c c c c}
  \toprule
  \multirow{2}{*}{TFE} & \multirow{2}{*}{MCP} & \multirow{2}{*}{DSSA} & \multirow{2}{*}{SVD} & \multicolumn{2}{c}{Valence} & \multicolumn{2}{c}{Arousal} \\
  \cmidrule(lr){5-6} \cmidrule(lr){7-8}
   &  &  &  & ACC & F1 & ACC & F1 \\
  \midrule
    &   &   &   & 65.83$\pm$12.34 & 56.84$\pm$17.26 & 60.83$\pm$12.76 & 66.43$\pm$16.42 \\
  $\checkmark$ &   &   &   & 63.75$\pm$11.18 & 61.17$\pm$14.37 & 62.92$\pm$10.86 & 65.17$\pm$15.28 \\
  $\checkmark$ & $\checkmark$ &   &   & 67.08$\pm$10.68 & 67.22$\pm$15.94 & 65.42$\pm$11.43 & 65.84$\pm$12.91 \\
  $\checkmark$ & $\checkmark$ & $\checkmark$ &   & 67.50$\pm$9.76 & 66.94$\pm$16.13 & 65.00$\pm$12.54 & 67.38$\pm$13.62 \\
  $\checkmark$ & $\checkmark$ & $\checkmark$ & $\checkmark$ & \textbf{67.92$\pm$8.45} & \textbf{66.67$\pm$13.45} & \textbf{65.42$\pm$9.73} & \textbf{68.44$\pm$12.45} \\
  \bottomrule
  \end{tabular}
  }
\end{table}
  
\subsubsection{Ablation on Loss Functions}
\begin{figure}[htbp]
  \centering
  \includegraphics[width=1.0\linewidth]{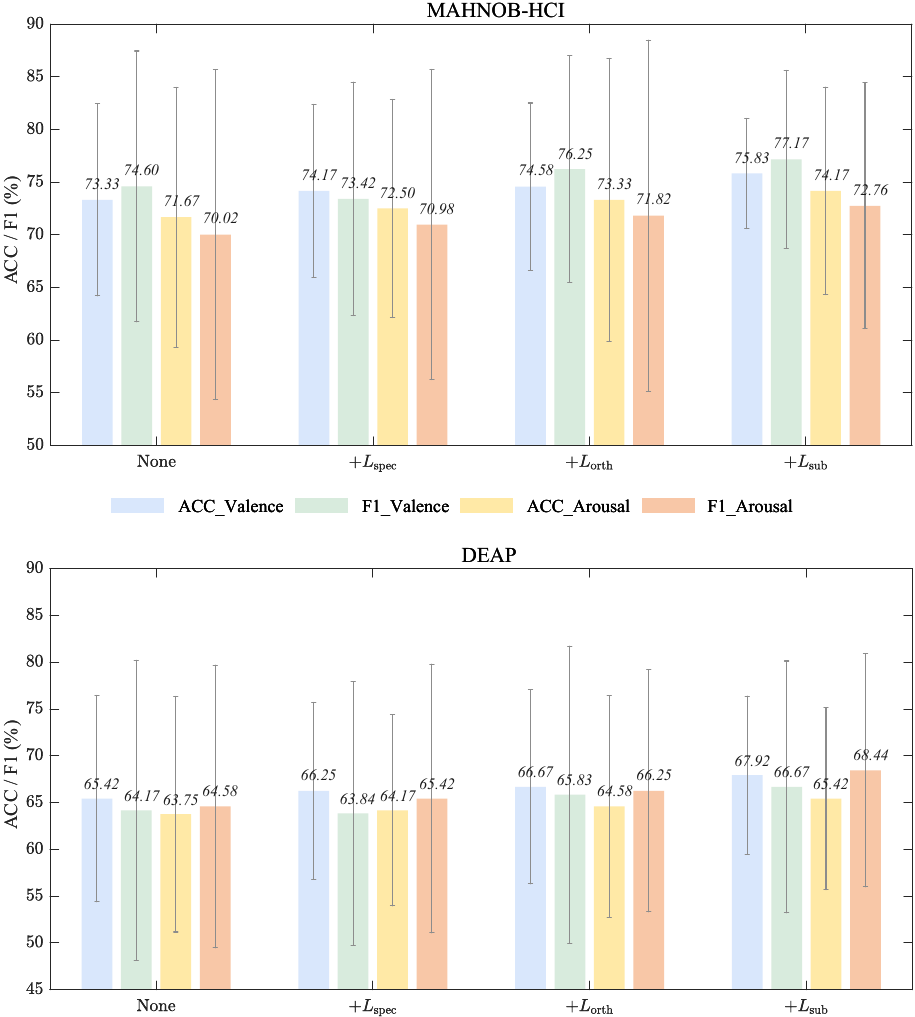}
  \caption{Ablation results on loss functions on MAHNOB-HCI (top) and DEAP (bottom). $L_{\mathrm{specific}}$, $L_{\mathrm{orth}}$, and $L_{\mathrm{sub}}$ are given in \eqref{eq:specific}, \eqref{eq:orth}, and \eqref{eq:sub}, respectively.}
  \label{fig:loss_ablation}
\end{figure}

We further ablate the loss design to examine the roles of sparsity regularization ($L_{\text{specific}}$), orthogonality regularization ($L_{\text{orth}}$), and subject supervision ($L_{\text{sub}}$). As shown in Fig.~\ref{fig:loss_ablation}, stacking these terms progressively improves or balances performance on both MAHNOB-HCI and DEAP, and the full objective consistently attains the strongest overall configuration. This pattern supports the view that the three losses are complementary for stabilizing shared-specific decoupling rather than substitutable. Qualitatively, $L_{\text{specific}}$ chiefly helps suppress noisy activations in the specific branch and tends to lift accuracy; $L_{\text{orth}}$ contributes more clearly to F1, especially on DEAP, consistent with reduced overlap between shared and specific corrections; and $L_{\text{sub}}$ yields the most pronounced gains on Arousal F1, indicating that explicit subject supervision anchors identity-related variation in the specific pathway and mitigates its leakage into the shared, emotion-relevant representation.

\begin{table*}[htbp]
  \centering
  \caption{Ablation results on different modality encoders.}
  \label{tab:encoder_ablation}
  \resizebox{0.95\textwidth}{!}{
  \begin{tabular}{c c c c c c c c c c}
  \toprule
  \multicolumn{2}{c}{\multirow{3}{*}{Model}} & \multicolumn{4}{c}{MAHNOB-HCI} & \multicolumn{4}{c}{DEAP} \\
  \cmidrule(lr){3-6} \cmidrule(lr){7-10}
  & & \multicolumn{2}{c}{Valence} & \multicolumn{2}{c}{Arousal} & \multicolumn{2}{c}{Valence} & \multicolumn{2}{c}{Arousal} \\
  \cmidrule(lr){3-4} \cmidrule(lr){5-6} \cmidrule(lr){7-8} \cmidrule(lr){9-10}
   & & ACC & F1 & ACC & F1 & ACC & F1 & ACC & F1 \\
  \midrule
  \multirow{4}{*}{\makecell{Different\\FE Encoder}}
  & ResNet-18 \cite{liu2023expression} & 65.83$\pm$12.84 & 68.70$\pm$16.72 & 68.33$\pm$10.93 & 66.63$\pm$17.86 & 65.83$\pm$13.58 & 53.93$\pm$19.24 & 60.42$\pm$11.47 & 67.13$\pm$16.91 \\
  & ResNet-50 \cite{zhang2021con} & 67.50$\pm$8.67 & 64.86$\pm$12.05 & 65.00$\pm$13.42 & 64.52$\pm$15.37 & 60.83$\pm$10.14 & 53.92$\pm$14.88 & 55.83$\pm$13.96 & 64.90$\pm$17.42 \\
  & I3D \cite{jiang2020dfew} & 68.33$\pm$11.26 & 72.46$\pm$14.91 & 66.67$\pm$9.38 & 69.85$\pm$16.74 & 56.67$\pm$12.67 & 67.30$\pm$18.35 & 56.67$\pm$10.92 & 71.43$\pm$14.76 \\
  & ViTB/16 \cite{chen2024static} & 71.67$\pm$9.84 & 76.39$\pm$13.18 & 70.00$\pm$12.47 & 68.48$\pm$15.42 & 65.83$\pm$12.34 & 56.84$\pm$17.26 & 60.83$\pm$12.76 & 66.43$\pm$16.42 \\
  \midrule
  \multirow{3}{*}{\makecell{Different\\rPPG Encoder}}
  & TCMA \cite{li2024end} & 55.83$\pm$11.83 & 49.47$\pm$15.64 & 63.33$\pm$12.47 & 61.92$\pm$15.29 & 54.17$\pm$12.41 & 60.85$\pm$18.57 & 52.50$\pm$13.78 & 59.92$\pm$16.38 \\
  & UBVMT \cite{ali2025unified} & 58.33$\pm$10.67 & 52.53$\pm$14.58 & 65.83$\pm$11.43 & 64.18$\pm$14.23 & 58.33$\pm$11.54 & 63.74$\pm$16.43 & 55.00$\pm$12.81 & 62.83$\pm$15.27 \\
  & TFE & 60.83$\pm$9.47 & 58.37$\pm$12.31 & 68.33$\pm$10.68 & 66.74$\pm$13.24 & 60.42$\pm$11.37 & 65.41$\pm$16.74 & 56.67$\pm$11.83 & 64.52$\pm$14.53 \\
  \midrule
  SCPT (Ours) & \makecell{ViTB/16\\+ TFE} & \textbf{75.83$\pm$5.24} & \textbf{77.17$\pm$8.45} & \textbf{74.17$\pm$9.85} & \textbf{72.76$\pm$11.67} & \textbf{67.92$\pm$8.45} & \textbf{66.67$\pm$13.45} & \textbf{65.42$\pm$9.73} & \textbf{68.44$\pm$12.45} \\
  \bottomrule
  \end{tabular}
  }
  \end{table*}

\subsubsection{Ablation on Different Modality Encoders}
To further examine the effect of unimodal feature quality, we compare different facial and physiological encoders in Table \ref{tab:encoder_ablation}. This analysis helps distinguish the benefit of improved unimodal representation learning from that of the proposed cross-modal interaction framework.

It can be observed from Table \ref{tab:encoder_ablation} that, among the facial encoders, the ViT-based backbone achieves the strongest overall performance, especially on MAHNOB-HCI. For example, ViTB/16 reaches 71.67$\pm$9.84 and 76.39$\pm$13.18 in Valence ACC/F1, together with 70.00$\pm$12.47 and 68.48$\pm$15.42 in Arousal ACC/F1, which are generally higher than the CNN-based alternatives. This result suggests that patch-wise tokenization and global self-attention are more effective than CNN-based alternatives in preserving subtle affective facial patterns for subsequent multimodal interaction. Among the rPPG encoders, TFE consistently outperforms TCMA and UBVMT on both datasets for both Valence and Arousal, indicating that stable time--frequency modeling is beneficial for extracting rhythm-related physiological cues. This advantage is particularly evident on MAHNOB-HCI, where TFE achieves 60.83$\pm$9.47 and 58.37$\pm$12.31 in Valence ACC/F1 and 68.33$\pm$10.68 and 66.74$\pm$13.24 in Arousal ACC/F1, and it remains competitive on DEAP despite the more challenging cross-subject setting.

It can also be observed that the full model maintains a clear performance margin over the encoder-only variants. These results indicate that the final gains cannot be attributed to encoder replacement alone, but rather to the joint effect of stronger unimodal representations, fine-grained prompt-guided cross-modal interaction, and subject-invariant adaptation within the proposed framework.

\begin{figure*}[htbp]
  \centering
  \includegraphics[width=1.0\linewidth]{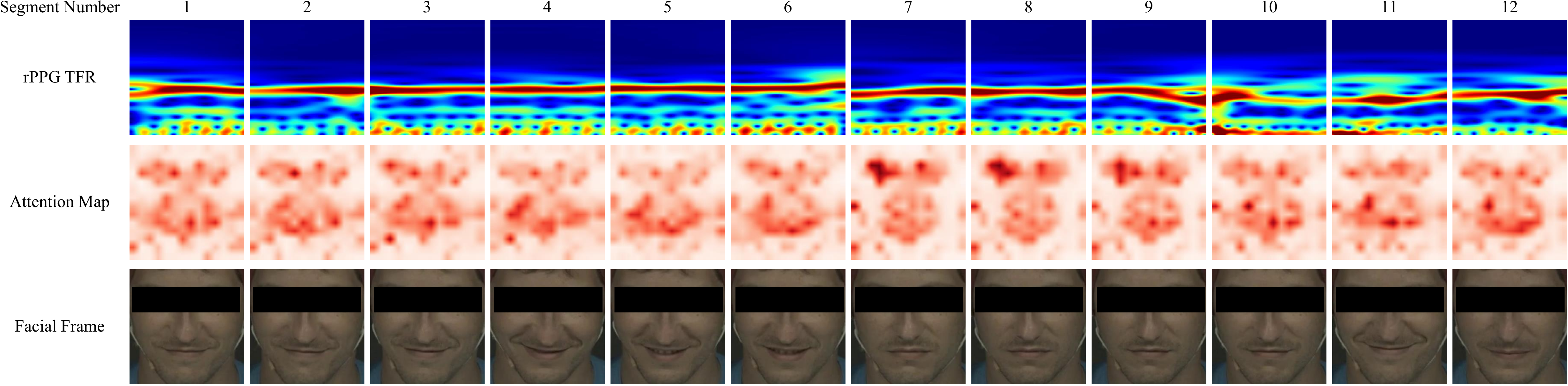}
  \caption{Temporal visualization of facial frames, rPPG time-frequency representations (TFRs), and spatial attention maps from the experimental segments of Subject-01 on the MAHNOB-HCI dataset. Top: rPPG TFRs. Middle: attention maps over the face region produced by the ViT backbone at each time step. Bottom: cropped facial frames. Columns follow chronological order within the clip, illustrating co-evolving pulse-related frequency structure and the facial regions emphasized by attention.}
  \label{fig:face_tfr_attention}
  \end{figure*}

\subsection{Visualization}
\begin{figure}[htbp]
  \centering
  \includegraphics[width=1.0\linewidth]{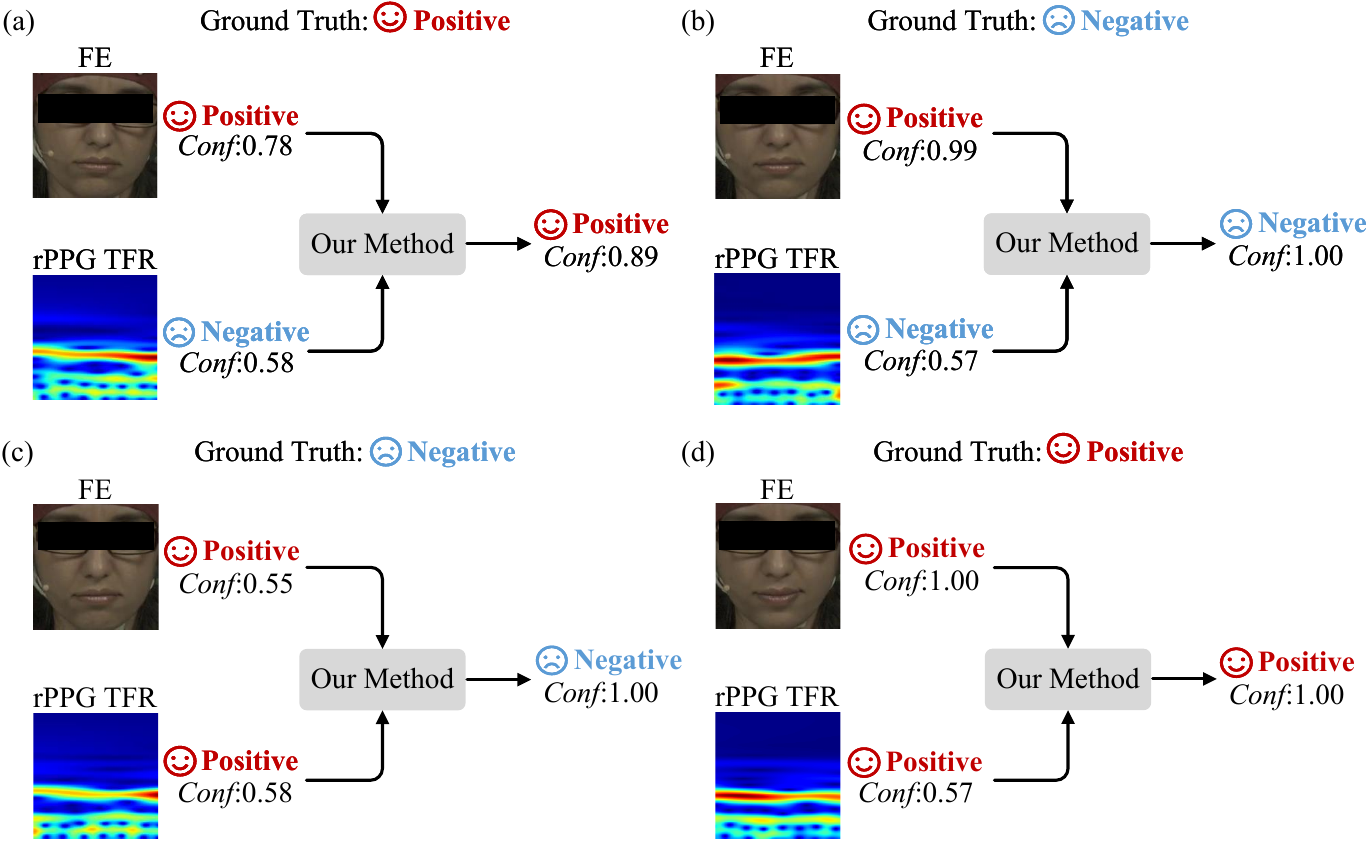}
  \caption{Several samples in MAHNOB-HCI dataset with inconsistent or consistent prediction on different modalities. Each panel shows the unimodal predictions from FE and rPPG alongside the fused prediction, with confidence scores (\textit{Conf}).}
  \label{fig:complementary_fusion}
  \end{figure}

Beyond the quantitative ablation results, we provide visual analyses on MAHNOB-HCI, including visualization of spatial attention maps (Fig.~\ref{fig:face_tfr_attention}), cross-modal complementary fusion examples (Fig.~\ref{fig:complementary_fusion}), t-SNE of the shared features under DSSA (Fig.~\ref{fig:tsne}), and the role of SVD in the shared subspace (Fig.~\ref{fig:svd_subspace}).

\subsubsection{Attention Visualization}
Fig.~\ref{fig:face_tfr_attention} illustrates the temporal evolution of facial expressions, rPPG TFRs, and spatial attention in a trial segment of Subject-01 from the MAHNOB-HCI dataset. Overall, the rPPG TFR maintains a stable periodic structure across segments, with localized energy enhancement observed around segments 7–10, reflecting dynamic variations in physiological rhythms. Correspondingly, the attention maps predominantly focus on rPPG-sensitive regions, such as the cheeks, nasal area, and perioral regions, indicating that the model can effectively identify spatial locations associated with blood flow variations. From the facial frames, it can be observed that as expressions transition from a relatively neutral state to more pronounced changes (e.g., slight smiling or muscle tension), the attention distribution adjusts accordingly, exhibiting a consistent dynamic response. This suggests that the model effectively exploits the complementary information of physiological and appearance cues during fusion, incorporating rPPG features while preserving facial expression semantics, thereby enhancing the discriminability and robustness of emotion recognition.

\subsubsection{Visualization of Cross-Modal Complementary Fusion}
To illustrate the complementary effect of cross-modal complementary fusion, we present several representative prediction examples from MAHNOB-HCI in Fig.~\ref{fig:complementary_fusion}. The results show that the two modalities provide complementary cues in both inconsistent and consistent prediction cases. By integrating FE and rPPG information through prompt-guided interaction, our method improves prediction confidence and corrects ambiguous or erroneous unimodal decisions. These examples demonstrate the effectiveness of MCP in exploiting cross-modal complementarity for more robust emotion recognition.

\subsubsection{Visualization of Shared Representation}
\begin{figure}[htbp]
  \centering
  \includegraphics[width=1.0\linewidth]{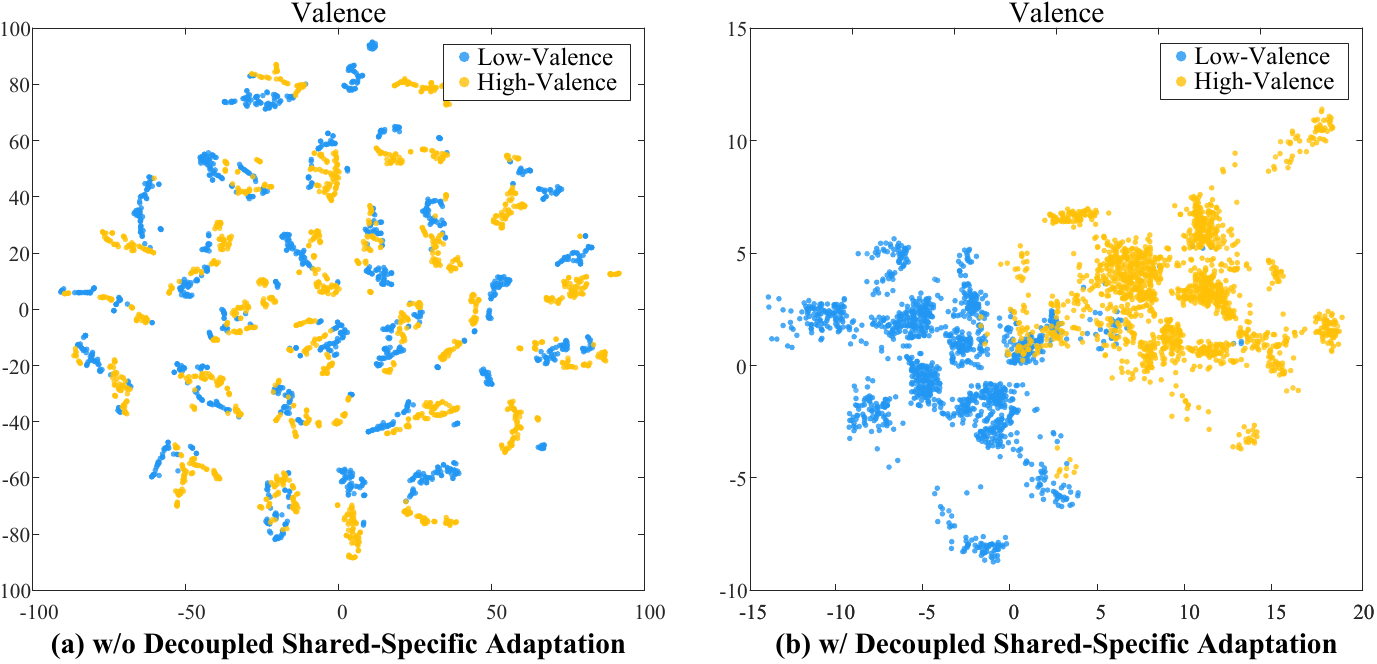}
  \\[0.4em]
  \includegraphics[width=1.0\linewidth]{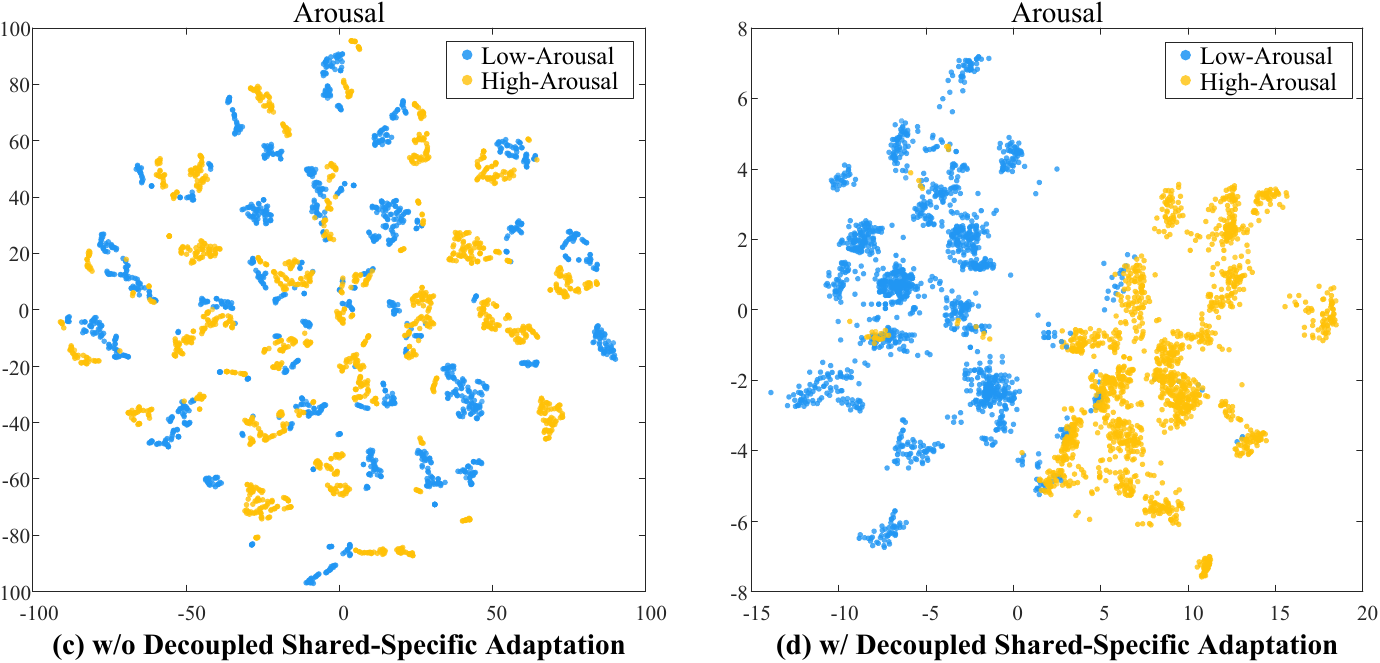}
  \caption{t-SNE visualization of the learned shared representations on MAHNOB-HCI.}
  \label{fig:tsne}
  \end{figure}

We further visualize the learned shared representations on MAHNOB-HCI using t-SNE to provide an intuitive view of the feature structure. As shown in Fig.~\ref{fig:tsne}, the representations learned without DSSA in Figs.~\ref{fig:tsne}(a) and \ref{fig:tsne}(c) are scattered over the embedding space, with low- and high-emotion samples exhibiting substantial local overlap and fragmented cluster structures. In contrast, after DSSA is introduced in Figs.~\ref{fig:tsne}(b) and \ref{fig:tsne}(d), the two classes become more compact and more clearly separated, especially along the dominant horizontal direction, indicating that the learned shared features are more discriminative for both Valence and Arousal.

The reduced overlap after DSSA is particularly notable around the decision boundary region, where many mixed samples in the w/o DSSA setting become reorganized into more coherent class-specific groups. This phenomenon suggests that DSSA helps suppress subject-dependent interference and reduces intra-class dispersion, thereby preserving emotion-relevant structure in the shared branch. These visual patterns are consistent with the quantitative gains obtained by DSSA in the ablation study, especially the more stable improvements on F1.

\subsubsection{Visualization of Singular Value Decomposition}
\begin{figure}[htbp]
\centering
\includegraphics[width=1.0\linewidth]{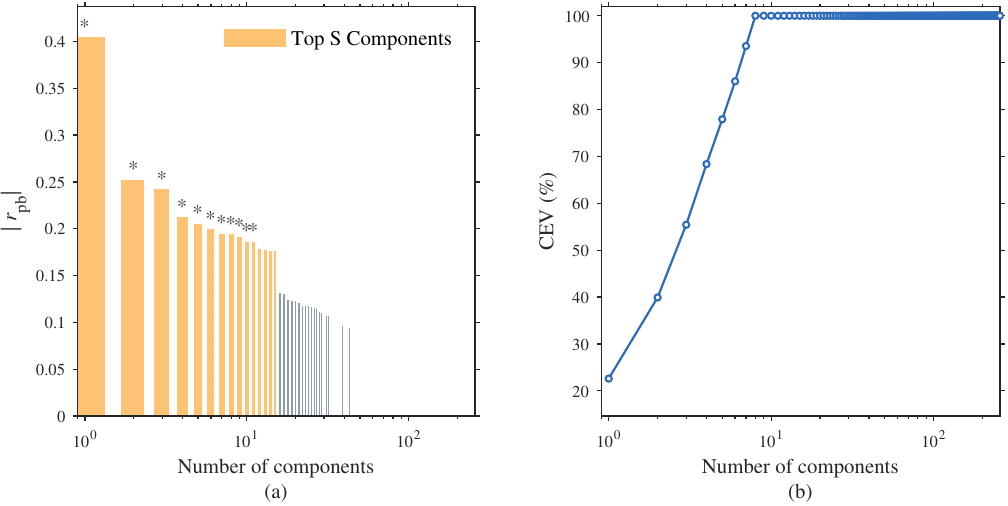}
\caption{Visualization of the role of SVD in the shared representation matrix on MAHNOB-HCI. (a): absolute point-biserial correlation $|r_{pb}|$ between affect labels and per-sample coordinates along each singular direction, with directions ranked by label relevance; the top $S$ directions are highlighted (orange) while the remaining directions are shown in gray. Gray asterisks mark statistically significant correlations at $p{<}0.05$. (b): cumulative explained variance (CEV) of the singular values in descending order.}
\label{fig:svd_subspace}
\end{figure}

To understand the role of SVD in the shared branch, we compute the absolute point-biserial correlation ($|r_{pb}|$) between affect labels and projection scores along each right singular vector of $\Gamma_{\mathrm{shared}}^{L}(X^{L})$ (Eq.~\ref{eq:vemo_derivation}), along with the cumulative explained variance (CEV) of the singular-value spectrum. As illustrated in Fig.~\ref{fig:svd_subspace}, (a)~ranks singular directions according to $|r_{pb}|$, showing a clear and rapid decay, with only a small subset of directions exhibiting strong label relevance; (b)~shows the CEV curve, which quickly approaches saturation within roughly ten components. This pattern indicates a steep singular-value spectrum, with most of the total variance explained by the first few singular directions, so the shared-branch matrix is energetically low rank and a compact basis suffices to preserve its dominant variability. These results provide intuitive support for the SVD-based design in Eq.~(\ref{eq:vemo_derivation}), suggesting that the primary emotion-related structure is concentrated in a low-dimensional subspace spanned by a few leading components.

\subsection{Sensitivity Analysis of Hyperparameters}
\begin{figure}[htbp]
\centering
\subfigure[MAHNOB-HCI]{\includegraphics[width=1.0\linewidth]{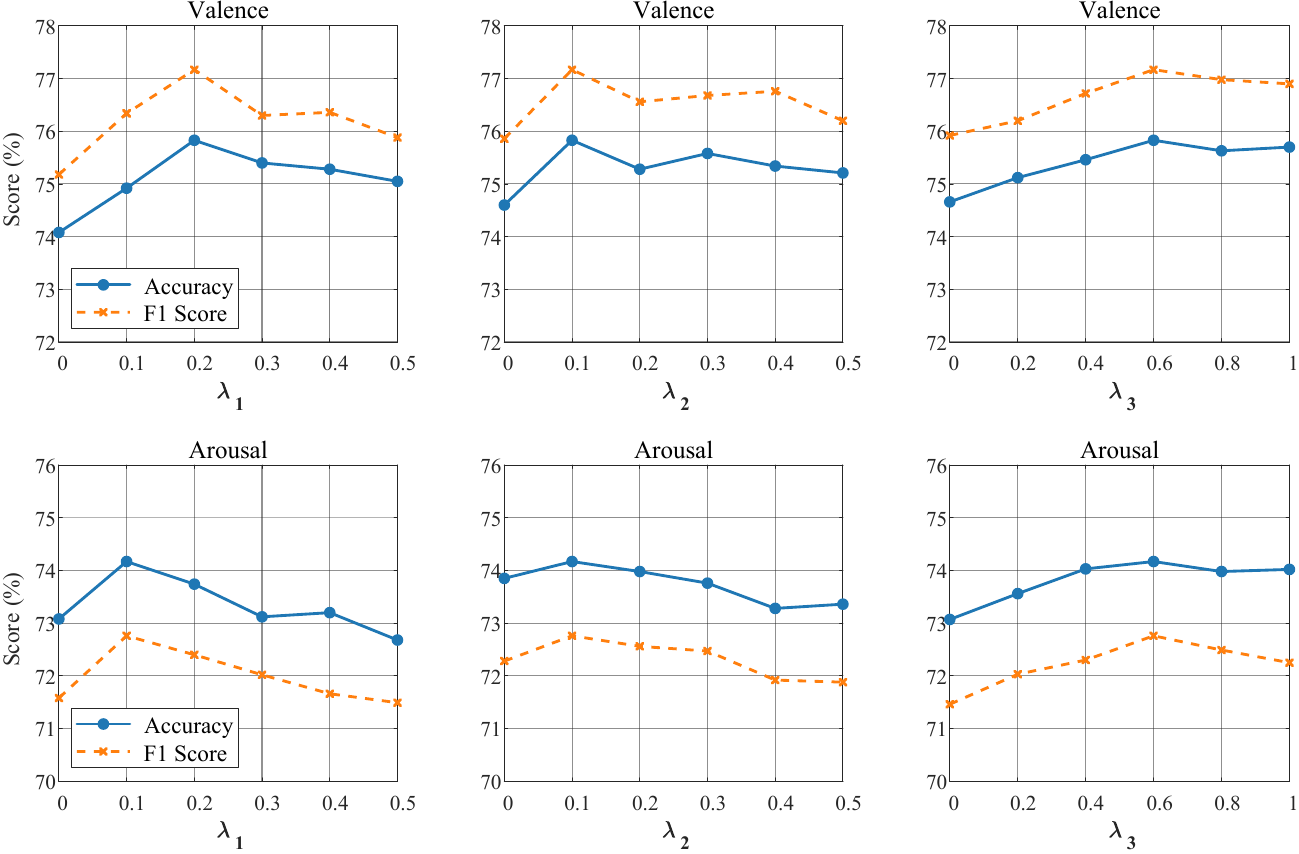}}
\subfigure[DEAP]{\includegraphics[width=1.0\linewidth]{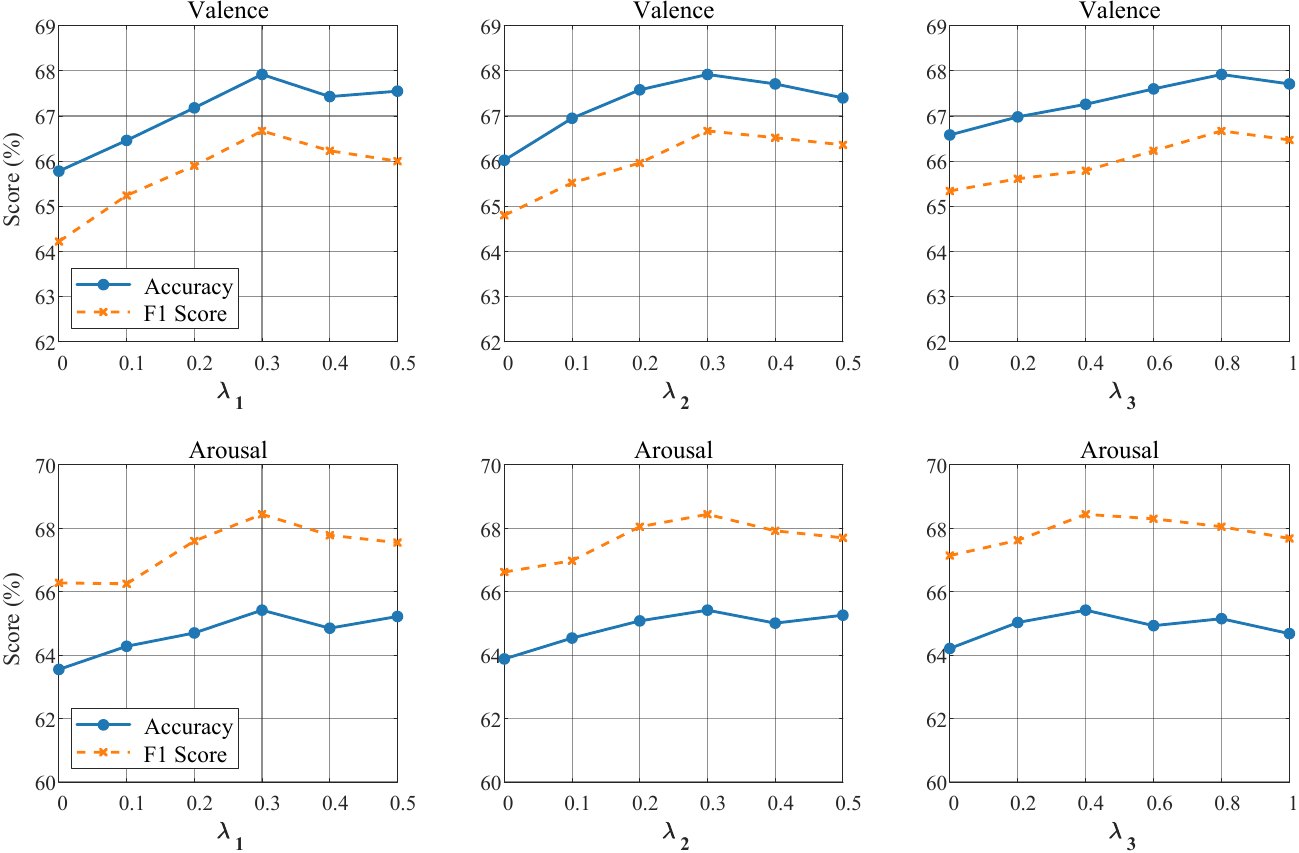}}
\caption{Sensitivity analysis of the weights $\lambda_1$, $\lambda_2$, and $\lambda_3$ in Eq.\eqref{eq:loss_dss} on MAHNOB-HCI and DEAP.}
\label{fig:lambda_sensitivity}
\end{figure}

Fig.~\ref{fig:lambda_sensitivity} illustrates the impact of auxiliary loss weights $(\lambda_1,\lambda_2,\lambda_3)$ on model performance across MAHNOB-HCI and DEAP. Overall, the performance varies smoothly with respect to each hyperparameter, indicating that the proposed framework is relatively robust to moderate weight changes. On MAHNOB-HCI, optimal performance is achieved when $\lambda_3$ is relatively large while $\lambda_1$ and $\lambda_2$ remain moderate, suggesting that emphasizing subject supervision effectively enhances emotion-discriminative representations. In contrast, DEAP exhibits better performance with slightly larger $\lambda_1$ and $\lambda_2$, reflecting the need for stronger sparsity and orthogonality constraints under more pronounced cross-subject variability. Notably, excessively large weights for any term lead to performance degradation, implying a trade-off between enforcing feature decoupling and preserving task-relevant information. These results demonstrate that the three auxiliary losses contribute complementarily, and balanced weighting is critical for achieving optimal performance.

\section{Conclusion}
In this paper, we propose SCPT, a subject-invariant cross-modal prompt-tuning framework for non-contact emotion recognition by jointly modeling facial expression representations and rPPG-based physical cues extracted from facial videos. The proposed framework tackles cross-modal complementary fusion and inter-subject variability via modality-complementary prompt injection and subject-invariant feature decoupling. Specifically, encoding rPPG signals in the time--frequency domain yields stable and discriminative physiological patterns that complement appearance-based facial features. The MCP module leverages these cues to enable fine-grained cross-modal interaction within the ViT backbone, resulting in more effective fusion than conventional coarse strategies. Meanwhile, the DSSA mechanism disentangles feature adaptation into subject-shared and subject-specific components, improving robustness to inter-subject variations and promoting more generalizable emotional representations. Extensive experiments on the MAHNOB-HCI and DEAP datasets demonstrate that SCPT consistently improves both recognition accuracy and cross-subject generalization. These results verify the effectiveness of combining physiologically grounded prompt-based interaction with decoupled shared-specific adaptation for robust multimodal emotion recognition. In future work, we will further investigate more flexible learning strategies for missing modalities and evaluate SCPT on larger and more diverse multimodal affective datasets.

\bibliographystyle{IEEEtran}
\bibliography{bibfile}

\begin{thebibliography}{10}
\providecommand{\url}[1]{#1}
\csname url@samestyle\endcsname
\providecommand{\newblock}{\relax}
\providecommand{\bibinfo}[2]{#2}
\providecommand{\BIBentrySTDinterwordspacing}{\spaceskip=0pt\relax}
\providecommand{\BIBentryALTinterwordstretchfactor}{4}
\providecommand{\BIBentryALTinterwordspacing}{\spaceskip=\fontdimen2\font plus
\BIBentryALTinterwordstretchfactor\fontdimen3\font minus
  \fontdimen4\font\relax}
\providecommand{\BIBforeignlanguage}[2]{{%
\expandafter\ifx\csname l@#1\endcsname\relax
\typeout{** WARNING: IEEEtran.bst: No hyphenation pattern has been}%
\typeout{** loaded for the language `#1'. Using the pattern for}%
\typeout{** the default language instead.}%
\else
\language=\csname l@#1\endcsname
\fi
#2}}
\providecommand{\BIBdecl}{\relax}
\BIBdecl

\bibitem{lerner2015emotion}
J.~S. Lerner, Y.~Li, P.~Valdesolo, and K.~S. Kassam, ``Emotion and decision
  making,'' \emph{Annu. Rev. Psychol.}, vol.~66, no.~1, pp. 799--823, 2015.

\bibitem{picard2000affective}
R.~W. Picard, \emph{Affective computing}.\hskip 1em plus 0.5em minus
  0.4em\relax MIT press, 2000.

\bibitem{xu2025hypercomplex}
X.~Xu, J.~Chen, C.~Fu, and Z.~Lyu, ``Hypercomplex neural network and
  cross-modal attention for multi-modal emotion recognition using physiological
  signals,'' \emph{IEEE Trans. Affect. Comput.}, vol.~16, no.~4, pp.
  3523--3536, 2025.

\bibitem{bamonte2023emotion}
M.~F. Bamonte, M.~Risk, and V.~Herrero, ``Emotion recognition based on galvanic
  skin response and photoplethysmography signals using artificial intelligence
  algorithms,'' in \emph{Congreso Argentino de Bioingenier{\'i}a}, 2023, pp.
  23--35.

\bibitem{song2020heart}
R.~Song, S.~Zhang, C.~Li, Y.~Zhang, J.~Cheng, and X.~Chen, ``Heart rate
  estimation from facial videos using a spatial-temporal representation with
  convolutional neural networks,'' \emph{IEEE Trans. Instrum. Meas.}, vol.~69,
  no.~10, pp. 7411--7421, 2020.

\bibitem{yu2022physformer}
Z.~Yu, Y.~Shen, J.~Shi, H.~Zhao, P.~Torr, and G.~Zhao, ``{PhysFormer}: Facial
  video-based physiological measurement with temporal difference transformer,''
  in \emph{Proc. IEEE/CVF Conf. Comput. Vis. Pattern Recognit. (CVPR)}, 2022,
  pp. 4186--4196.

\bibitem{cheng2023motion}
J.~Cheng, R.~Liu, J.~Li, R.~Song, Y.~Liu, and X.~Chen, ``Motion-robust
  respiratory rate estimation from camera videos via fusing pixel movement and
  pixel intensity information,'' \emph{IEEE Trans. Instrum. Meas.}, vol.~72,
  pp. 1--11, 2023.

\bibitem{liu2024rppg}
X.~Liu, Y.~Zhang, Z.~Yu, H.~Lu, H.~Yue, and J.~Yang, ``{rPPG-MAE}:
  Self-supervised pretraining with masked autoencoders for remote physiological
  measurements,'' \emph{IEEE Trans. Multimedia}, vol.~26, pp. 7278--7293, 2024.

\bibitem{zou2025rhythmmamba}
B.~Zou, Z.~Guo, X.~Hu, and H.~Ma, ``{RhythmMamba}: Fast, lightweight, and
  accurate remote physiological measurement,'' in \emph{Proc. AAAI Conf. Artif.
  Intell.}, vol.~39, no.~10, 2025, pp. 11\,077--11\,085.

\bibitem{benezeth2018remote}
Y.~Benezeth, P.~Li, R.~Macwan, K.~Nakamura, R.~Gomez, and F.~Yang, ``Remote
  heart rate variability for emotional state monitoring,'' in \emph{Proc. IEEE
  EMBS Int. Conf. Biomed. Health Inform. (BHI)}, 2018, pp. 1--4.

\bibitem{yu2019remote}
Z.~Yu, X.~Li, and G.~Zhao, ``Remote photoplethysmograph signal measurement from
  facial videos using spatio-temporal networks,'' in \emph{Proc. Brit. Mach.
  Vision Conf. (BMVC)}, 2019, pp. 1--12.

\bibitem{mellouk2023cnn}
W.~Mellouk and W.~Handouzi, ``{CNN-LSTM} for automatic emotion recognition
  using contactless photoplythesmographic signals,'' \emph{Biomed. Signal
  Process. Control}, vol.~85, p. 104907, 2023.

\bibitem{ouzar2022video}
Y.~Ouzar, F.~Bousefsaf, D.~Djeldjli, and C.~Maaoui, ``Video-based multimodal
  spontaneous emotion recognition using facial expressions and physiological
  signals,'' in \emph{Proc. IEEE/CVF Conf. Comput. Vis. Pattern Recognit.
  (CVPR)}, 2022, pp. 2459--2468.

\bibitem{wu2023recognizing}
Y.-C. Wu, L.-W. Chiu, C.-C. Lai, B.-F. Wu, and S.~S.~J. Lin, ``Recognizing,
  fast and slow: Complex emotion recognition with facial expression detection
  and remote physiological measurement,'' \emph{IEEE Trans. Affect. Comput.},
  vol.~14, no.~4, pp. 3177--3190, 2023.

\bibitem{li2024end}
J.~Li and J.~Peng, ``End-to-end multimodal emotion recognition based on facial
  expressions and remote photoplethysmography signals,'' \emph{IEEE J. Biomed.
  Health Inform.}, 2024.

\bibitem{bao2025svd}
J.~Bao, J.~Qian, and J.~Yang, ``{SVD}-guided multimodal feature fusion for
  emotion recognition from facial videos,'' \emph{IEEE Trans. Affect. Comput.},
  vol.~16, no.~3, pp. 1705--1715, 2025.

\bibitem{shen2023contrastive}
X.~Shen, X.~Liu, X.~Hu, D.~Zhang, and S.~Song, ``Contrastive learning of
  subject-invariant eeg representations for cross-subject emotion
  recognition,'' \emph{IEEE Transactions on Affective Computing}, vol.~14,
  no.~3, pp. 2496--2511, 2023.

\bibitem{wang2023generalizing}
J.~Wang, C.~Lan, C.~Liu, Y.~Ouyang, T.~Qin, W.~Lu, Y.~Chen, W.~Zeng, and P.~S.
  Yu, ``Generalizing to unseen domains: A survey on domain generalization,''
  \emph{IEEE Trans. Knowl. Data Eng.}, vol.~35, no.~8, pp. 8052--8072, 2023.

\bibitem{zhao2021plug}
L.-M. Zhao, X.~Yan, and B.-L. Lu, ``Plug-and-play domain adaptation for
  cross-subject {EEG}-based emotion recognition,'' in \emph{Proc. AAAI Conf.
  Artif. Intell. (AAAI)}, 2021, pp. 863--870.

\bibitem{jia2024multi}
Z.~Jia, F.~Zhao, Y.~Guo, H.~Chen, T.~Jiang, and B.~Center, ``Multi-level
  disentangling network for cross-subject emotion recognition based on
  multimodal physiological signals,'' in \emph{Proc. Int. Joint Conf. Artif.
  Intell. (IJCAI)}, 2024, pp. 3069--3077.

\bibitem{lyu2025mi}
Z.~Lyu, Z.~Zuo, C.~Chen, and Y.~Fang, ``Mutual information disentanglement
  based domain adaptation model for {EEG} emotion recognition,'' \emph{IEEE
  Signal Process. Lett.}, vol.~32, pp. 3027--3031, 2025.

\bibitem{yang2026fddgnet}
Y.~Yang, L.~Duan, K.~Hou, Z.~Kang, X.~Zhang, and B.~Hu, ``{FDDGNet}: An
  information bottleneck-inspired feature disentanglement network for
  cross-subject {EEG}-based emotion recognition,'' \emph{Neurocomputing}, vol.
  668, p. 132368, 2026.

\bibitem{gao2025multimodal}
H.~Gao, Z.~Cai, X.~Wang, M.~Wu, and C.~Liu, ``Multimodal fusion of behavioral
  and physiological signals for enhanced emotion recognition via feature
  decoupling and knowledge transfer,'' \emph{IEEE J. Biomed. Health Inform.},
  2025.

\bibitem{piratla2020efficient}
V.~Piratla, P.~Netrapalli, and S.~Sarawagi, ``Efficient domain generalization
  via common-specific low-rank decomposition,'' in \emph{Proc. Int. Conf. Mach.
  Learn. (ICML)}, 2020, pp. 7728--7738.

\bibitem{li2018deep}
Y.~Li, X.~Tian, M.~Gong, Y.~Liu, T.~Liu, K.~Zhang, and D.~Tao, ``Deep domain
  generalization via conditional invariant adversarial networks,'' \emph{Proc.
  Eur. Conf. Comput. Vis. (ECCV)}, pp. 624--639, 2018.

\bibitem{zhou2025eegmatch}
R.~Zhou, W.~Ye, Z.~Zhang, Y.~Luo, L.~Zhang, L.~Li, G.~Huang, Y.~Dong, Y.-T.
  Zhang, and Z.~Liang, ``Eegmatch: Learning with incomplete labels for
  semisupervised eeg-based cross-subject emotion recognition,'' \emph{IEEE
  Trans. Neural Netw. Learn. Syst.}, vol.~36, no.~7, pp. 12\,991--13\,005,
  2025.

\bibitem{li2025emotion}
J.~Li, J.~Nie, D.~Guo, R.~Hong, and M.~Wang, ``Emotion separation and
  recognition from a facial expression by generating the poker face with vision
  transformers,'' \emph{IEEE Transactions on Computational Social Systems},
  vol.~12, no.~4, pp. 1548--1562, 2025.

\bibitem{ismail2022comparison}
S.~N. M.~S. Ismail, N.~A.~A. Aziz, and S.~Z. Ibrahim, ``A comparison of emotion
  recognition system using electrocardiogram ({ECG}) and photoplethysmogram
  ({PPG}),'' \emph{J. King Saud Univ.-Comput. Inform. Sci.}, vol.~34, no.~6,
  pp. 3539--3558, 2022.

\bibitem{sarkar2020self}
P.~Sarkar and A.~Etemad, ``Self-supervised {ECG} representation learning for
  emotion recognition,'' \emph{IEEE Trans. Affect. Comput.}, vol.~13, no.~3,
  pp. 1541--1554, 2020.

\bibitem{zhu2024dynamic}
Q.~Zhu, C.~Zheng, Z.~Zhang, W.~Shao, and D.~Zhang, ``Dynamic confidence-aware
  multi-modal emotion recognition,'' \emph{IEEE Trans. Affect. Comput.},
  vol.~15, no.~3, pp. 1358--1370, 2024.

\bibitem{han2023noise}
J.~Han, X.~Gu, G.-Z. Yang, and B.~Lo, ``Noise-factorized disentangled
  representation learning for generalizable motor imagery {EEG}
  classification,'' \emph{IEEE J. Biomed. Health Inform.}, vol.~28, no.~2, pp.
  765--776, 2023.

\bibitem{wu2024grop}
M.~Wu, C.~P. Chen, B.~Chen, and T.~Zhang, ``{Grop}: Graph orthogonal
  purification network for {EEG} emotion recognition,'' \emph{IEEE Trans.
  Affect. Comput.}, 2024.

\bibitem{cheng2026video}
J.~Cheng, X.~Luo, X.~Wu, R.~Song, and Y.~Liu, ``Video-based instantaneous heart
  rate measurement with enhanced time-frequency representations,'' \emph{IEEE
  Transactions on Multimedia}, vol.~28, pp. 1289--1301, 2026.

\bibitem{mollahosseini2017affectnet}
A.~Mollahosseini, B.~Hasani, and M.~H. Mahoor, ``{AffectNet}: A database for
  facial expression, valence, and arousal computing in the wild,'' \emph{IEEE
  Trans. Affect. Comput.}, vol.~10, no.~1, pp. 18--31, 2017.

\bibitem{chen2024static}
Y.~Chen, J.~Li, S.~Shan, M.~Wang, and R.~Hong, ``From static to dynamic:
  Adapting landmark-aware image models for facial expression recognition in
  videos,'' \emph{IEEE Trans. Affect. Comput.}, vol.~16, no.~2, pp. 624--638,
  2024.

\bibitem{zhu2023visual}
J.~Zhu, S.~Lai, X.~Chen, D.~Wang, and H.~Lu, ``Visual prompt multi-modal
  tracking,'' in \emph{Proc. IEEE/CVF Conf. Comput. Vis. Pattern Recognit.
  (CVPR)}, 2023, pp. 9516--9526.

\bibitem{hsieh2026alfa}
H.-Y. Hsieh, W.-T.~M. Ting, and H.~T. Kung, ``{Alfa}: Attentive low-rank filter
  adaptation for structure-aware cross-domain personalized gaze estimation,''
  in \emph{Proc. AAAI Conf. Artif. Intell. (AAAI)}, 2026, pp. 17\,481--17\,489.

\bibitem{wei2001ecg}
J.-J. Wei, C.-J. Chang, N.-K. Chou, and G.-J. Jan, ``{ECG} data compression
  using truncated singular value decomposition,'' \emph{IEEE Trans. Inf.
  Technol. Biomed.}, vol.~5, no.~4, pp. 290--299, 2001.

\bibitem{ferdinando2017emotion}
H.~Ferdinando, T.~Sepp{\"a}nen, and E.~Alasaarela, ``Emotion recognition using
  neighborhood components analysis and {ECG/HRV}-based features,'' in
  \emph{Proc. Int. Conf. Pattern Recognit. Appl. Methods (ICPRAM)}, 2017, pp.
  99--113.

\bibitem{oguz2023emotion}
F.~E. Oğuz, A.~Alkan, and T.~Schöler, ``Emotion detection from {ECG} signals
  with different learning algorithms and automated feature engineering,''
  \emph{Signal Image Video Process.}, vol.~17, no.~7, pp. 3783--3791, 2023.

\bibitem{singh2023bio}
A.~Singh, T.~Wittenberg, M.-M. Salman, N.~Holzer, S.~Göb, J.~Pahl, T.~Götz,
  and S.~Sawant, ``Bio-signal based multimodal fusion with bilinear model for
  emotion recognition,'' in \emph{2023 IEEE International Conference on
  Bioinformatics and Biomedicine (BIBM)}, 2023, pp. 4834--4839.

\bibitem{jung2019utilizing}
Siddharth, T.-P. Jung, and T.~J. Sejnowski, ``Utilizing deep learning towards
  multi-modal bio-sensing and vision-based affective computing,'' \emph{IEEE
  Trans. Affect. Comput.}, vol.~13, no.~1, pp. 96--107, 2019.

\bibitem{li2021mindlink}
R.~Li \emph{et~al.}, ``{MindLink-eumpy}: an open-source python toolbox for
  multimodal emotion recognition,'' \emph{Front. Hum. Neurosci.}, vol.~15, p.
  621493, 2021.

\bibitem{tan2021neurosense}
C.~Tan, M.~Šarlija, and N.~Kasabov, ``{NeuroSense}: Short-term emotion
  recognition and understanding based on spiking neural network modelling of
  spatio-temporal {EEG} patterns,'' \emph{Neurocomputing}, vol. 434, pp.
  137--148, 2021.

\bibitem{ali2025unified}
K.~Ali and C.~E. Hughes, ``A unified biosensor--vision multi-modal transformer
  network for emotion recognition,'' \emph{Biomed. Signal Process. Control},
  vol. 102, p. 107232, 2025.

\bibitem{wu2026dhcm}
B.~Wu and Y.~Li, ``{DHCM-CACL}: Dynamic hierarchical cross-modal mamba with
  confidence-adaptive contrastive learning for multimodal emotion
  recognition,'' in \emph{Proc. AAAI Conf. Artif. Intell. (AAAI)}, 2026, pp.
  2164--2172.

\bibitem{soleymani2011multimodal}
M.~Soleymani, J.~Lichtenauer, T.~Pun, and M.~Pantic, ``A multimodal database
  for affect recognition and implicit tagging,'' \emph{IEEE Trans. Affect.
  Comput.}, vol.~3, no.~1, pp. 42--55, 2012.

\bibitem{koelstra2012deap}
S.~Koelstra, C.~Muhl, M.~Soleymani, J.-S. Lee, A.~Yazdani, T.~Ebrahimi, T.~Pun,
  A.~Nijholt, and I.~Patras, ``{DEAP}: A database for emotion analysis using
  physiological signals,'' \emph{IEEE Trans. Affect. Comput.}, vol.~3, no.~1,
  pp. 18--31, 2012.

\bibitem{lugaresi2019mediapipe}
C.~Lugaresi \emph{et~al.}, ``{MediaPipe}: A framework for building perception
  pipelines,'' \emph{arXiv preprint arXiv:1906.08172}, 2019.

\bibitem{he2016identity}
K.~He, X.~Zhang, S.~Ren, and J.~Sun, ``Identity mappings in deep residual
  networks,'' in \emph{Proc. Eur. Conf. Comput. Vis. (ECCV)}, 2016, pp.
  630--645.

\bibitem{zhang2022gcb}
T.~Zhang, X.~Wang, X.~Xu, and C.~P. Chen, ``{GCB-Net}: Graph convolutional
  broad network and its application in emotion recognition,'' \emph{IEEE Trans.
  Affect. Comput.}, vol.~13, no.~1, pp. 379--388, 2022.

\bibitem{elalamy2021multi}
R.~Elalamy, M.~Fanourakis, and G.~Chanel, ``Multi-modal emotion recognition
  using recurrence plots and transfer learning on physiological signals,'' in
  \emph{Proc. Int. Conf. Affect. Comput. Intell. Interact. (ACII)}, 2021, pp.
  1--7.

\bibitem{li2025uncertainty}
G.~Li, N.~Chen, H.~Zhu, J.~Li, Z.~Xu, and Z.~Zhu, ``Uncertainty-aware graph
  contrastive fusion network for multimodal physiological signal emotion
  recognition,'' \emph{Neural Netw.}, vol. 187, p. 107363, 2025.

\bibitem{tian2026heterogeneity}
Y.~Tian, J.~Li, N.~Chen, G.~Li, Z.~Xu, H.~Zhu, Y.~Li, and Z.~Zhu,
  ``Heterogeneity-aware multi-modal physiological signal fusion strategy based
  on combined contrastive learning for emotion recognition,'' \emph{Neural
  Networks}, p. 108818, 2026.

\bibitem{liu2023expression}
Y.~Liu, W.~Wang, C.~Feng, H.~Zhang, Z.~Chen, and Y.~Zhan, ``Expression snippet
  transformer for robust video-based facial expression recognition,''
  \emph{Pattern Recognit.}, vol. 138, p. 109368, 2023.

\bibitem{zhang2021con}
S.~Zhang, Y.~Ding, Z.~Wei, and C.~Guan, ``Continuous emotion recognition with
  audio-visual leader-follower attentive fusion,'' in \emph{Proc. IEEE/CVF Int.
  Conf. Comput. Vis. Workshops (ICCVW)}, 2021, pp. 3560--3567.

\bibitem{jiang2020dfew}
X.~Jiang, Y.~Zong, W.~Zheng, C.~Tang, W.~Xia, C.~Lu, and J.~Liu, ``{DFEW}: A
  large-scale database for recognizing dynamic facial expressions in the
  wild,'' in \emph{Proc. 28th ACM Int. Conf. Multimedia (MM)}, 2020, pp.
  2881--2889.

\end{thebibliography}

\end{document}